\documentclass[pdflatex,sn-chicago]{sn-jnl}

\usepackage{graphicx}%
\usepackage{multirow}%
\usepackage{amsmath,amssymb,amsfonts}%
\usepackage{amsthm}%
\usepackage{mathrsfs}%
\usepackage[title]{appendix}%
\usepackage{xcolor}%
\usepackage{textcomp}%
\usepackage{manyfoot}%
\usepackage{booktabs}%
\usepackage{algorithm}%
\usepackage{algorithmicx}%
\usepackage{algpseudocode}%
\usepackage{listings}%

\theoremstyle{thmstyleone}%

\theoremstyle{thmstyletwo}%

\theoremstyle{thmstylethree}%

\begin{document}

\title[Article Title]{Parameter-Efficient Fine-Tuning of Multispectral Foundation Models for Hyperspectral Image Classification}


\author*[1]{\fnm{Bernardin} \sur{LIGAN}}\email{bernardin.ligan@um6p.ma}

\author[1,2]{\fnm{Khalide} \sur{Jbilou}}\email{khalide.jbilou@univ-littoral.fr}
\equalcont{These authors contributed equally to this work.}

\author[1,3]{\fnm{Fahd} \sur{Kalloubi}}\email{fahd.kalloubi@gmail.com}
\equalcont{These authors contributed equally to this work.}

\author[1]{\fnm{Ahmed} \sur{Ratnani}}\email{ahmed.ratnani@um6p.ma}
\equalcont{These authors contributed equally to this work.}

\affil*[1]{\orgdiv{The UM6P Vanguard Center}, \orgname{University Mohammed VI Polytechnic (UM6P)}, 
\orgaddress{
\city{Ben Guerir}, 
\country{Morocco}}}

\affil[2]{\orgdiv{Laboratoire de Mathématiques Pures et Appliquées
(LMPA)}, \orgname{Université du Littoral Cote d'Opale (ULCO)}, \orgaddress{ \city{Calais}, 
\country{France}}}

\affil[3]{\orgdiv{Faculty of Sciences Semlalia}, \orgname{Cadi Ayyad University}, \orgaddress{
\city{Marrakech}, 
\country{Morocco}}}


\abstract{Foundation models have experienced tremendous success across diverse fields, including remote sensing (RS), due to their versatility and strong generalization capabilities. However, most foundation models in RS are designed for multispectral imaging, with limited efforts devoted to hyperspectral imagery, which presents unique challenges due to its hundreds of spectral bands. Moreover, the efficient fine-tuning of foundation models for downstream tasks remains a challenging issue, as it often requires substantial memory and storage resources. 
In this paper, we present an effective framework to fine-tune SpectralGPT, a multispectral foundation model, for hyperspectral image classification (HSIC) tasks. To improve its efficiency, we explore various well-established Parameter-Efficient Fine-Tuning (PEFT) methods, including Low-Rank Adaptation (LoRA), Kronecker product-based adaptation (KronA), Low-Rank Kronecker product adaptation (LoKr), and the recent LoRA+, an extension of LoRA that employs distinct learning rates, linked by a scaling factor $\lambda$, for each low-rank adapter matrix during training. Inspired by LoRA+, we propose KronA+, an extension of KronA that applies a similar mechanism to the matrices used in the Kronecker product. We evaluate our methods on five datasets from different sensors, demonstrating competitive performance with state-of-the-art backbones in the hyperspectral domain. Our full fine-tuning (FFT) framework for SpectralGPT even surpasses a dedicated hyperspectral foundation model on certain datasets while requiring only about a quarter of the training epochs. Under identical training epochs and with significantly fewer trainable parameters compared to FFT of SpectralGPT, KronA+ achieves  comparable performance, demonstrating its effectiveness as the best-performing PEFT method. It utilizes only 0.056\% of trainable parameters and adds merely $\sim$0.2 MB of storage overhead.  
\footnote{This work is currently under review.}
}




\maketitle

\section{Introduction}\label{sec1}

Hyperspectral imaging (HSI) has emerged as a powerful technology for applications requiring precise spectral characterization of materials, enabling advancements in environmental monitoring, agriculture, defense, and mineral exploration. Unlike traditional RGB or multispectral data, which capture a limited number of spectral bands, HSI provides a continuous spectrum for each pixel across hundreds of narrow wavelengths. 
The unique ability of HSI to capture subtle spectral differences is critical for applications where traditional imaging methods lack the necessary depth. However, harnessing the full potential of HSI data presents significant challenges. Its high dimensionality and complex structure require advanced machine learning and deep learning models capable of extracting meaningful patterns. Despite progress in this area, performance gaps remain, particularly in classification tasks where distinguishing fine spectral variations is essential.

Foundation models have revolutionized natural language processing (NLP) by leveraging large-scale datasets to generalize across diverse tasks. Models such as BERT \citep{bert} and GPT \citep{gpt} are pre-trained on vast corpora and fine-tuned to adapt to a variety of downstream applications. This paradigm shift has led to significant advancements in NLP and has been successfully extended to other domains, notably computer vision. In computer vision, foundation models like Segment Anything Model (SAM) \citep{sam}, and CLIP \citep{clip} are pre-trained on large-scale RGB image datasets and achieve impressive performance across a range of tasks, including image classification, object detection, and segmentation.

Encouraged by the success of foundation models in NLP and computer vision, researchers have begun to develop such models for remote sensing tasks \citep{fmsreview}. Most of these models, however, are trained primarily on multispectral data, which has historically been more accessible and widely utilized. To our knowledge, only a few foundation models have been specifically designed for hyperspectral data. These hyperspectral foundation models, while promising, are typically more complex and computationally expensive, as they must process the high-dimensional spatial-spectral information inherent to hyperspectral imagery.
Fine-tuning large-scale models for specific tasks presents significant challenges, particularly due to the high computational and storage demands. As foundation models continue to grow in scale, the development of efficient fine-tuning methods becomes increasingly critical, especially in resource-constrained environments.

To address these challenges, Parameter-Efficient Fine-Tuning (PEFT) methods, such as LoRA \citep{lora}, have emerged. LoRA (Low-Rank Adaptation) introduces low-rank decomposition into the model's weight updates during fine-tuning, enabling the adaptation of large models to downstream tasks with only a small subset of parameters. Alternative techniques to LoRA such as KronA\citep{krona} and Lokr\citep{lycoris}, have been proposed to address the limitations of low-rank constraints in LoRA, enhancing expressiveness while maintaining efficiency. Recent work on LoRA+  \citep{lora+} has demonstrated that using distinct learning rates for low-rank matrices can further improve fine-tuning performance.

Inspired by the advancements in PEFT methods and the limited availability of hyperspectral-specific foundation models, two key questions arise:
\begin{enumerate}
    \item Rather than developing a new hyperspectral-specific foundation model, why not fine-tune an existing multispectral foundation model for hyperspectral image classification? 
    \item How can parameter-efficient fine-tuning (PEFT) methods be leveraged to reduce the number of trainable parameters during this adaptation while maintaining performance comparable to full fine-tuning?
\end{enumerate}

Building on these questions, our contributions can be summarized as follows:

\begin{itemize}
\item We introduce the first attempt to fine-tune a multispectral foundation model (SpectralGPT) for hyperspectral image classification, demonstrating that, with appropriate dataset transformations and fine-tuning strategies, it can achieve performance comparable to dedicated hyperspectral architectures.
\item We explore a range of PEFT techniques, including LoRA, LoRA+, KronA, Lokr, and introduce KronA+—a novel extension of KronA inspired by LoRA+ that applies differentiated learning rates to the matrices involved in the Kronecker product. This strategy significantly reduces the number of trainable parameters while maintaining performance competitive with full fine-tuning. To the best of our knowledge, this is the first systematic study on PEFT methods in hyperspectral image classification, offering new insights into the trade-offs between efficiency and performance in adapting foundation models to hyperspectral data.
\item We conduct a comprehensive evaluation of our framework on five diverse hyperspectral datasets, demonstrating its effectiveness across different sensors and imaging environments.
\end{itemize}

\footnote{This work is currently under review.}

\section{Related Work}
\subsection{Deep Learning for Hyperspectral Image Classification}

Hyperspectral images (HSI) are represented as 3D tensors $\textit{X} \in \mathbb{R}^{H \times W \times B}$, where $H$ and $W$ denote spatial dimensions, and $B$ is the number of spectral bands. The classification task involves assigning a label $y \in Y$ to each pixel based on its spectral signature. 
Beyond spectral features, spatial information is crucial in HSIC. To integrate spectral-spatial context, hyperspectral images are generally divided into patches $X_p \in \mathbb{R}^{P \times P \times B}$, with classification performed on the center pixel. Deep learning models, particularly convolutional neural networks (CNNs), have significantly improved HSIC by jointly learning spectral and spatial representations  \citep{patch1, spectralspatial1, spectralspatial2, spectralspatial3}. More recently, transformer-based models  \citep{spectralformer} have leveraged self-attention  \citep{attention1} to model long-range dependencies across spectral bands, overcoming CNN limitations. State-space models (SSMs) \citep{ssm}, such as MambaHSI  \citep{mambahsi}, further enhance classification by efficiently capturing long-range spatial and spectral relationships with linear computational complexity.

\subsection{Remote Sensing Foundation Models}
Foundation models in remote sensing have been primarily advanced through self-supervised learning, addressing the scarcity of labeled data. A widely used technique is Masked Image Modeling (MIM), often implemented via Masked Autoencoders (MAE)  \citep{mae1}, where a model learns meaningful representations by reconstructing masked image regions.

Several remote sensing models leverage this approach. SatMAE  \citep{satmae} specializes in temporal and multispectral satellite imagery, while SatMAE++  \citep{satmae++} improves multiscale representation learning. Prithvi  \citep{prithvi} enhances spatiotemporal modeling through 3D positional and patch embeddings in a Vision Transformer (ViT) framework. SpectralGPT  \citep{spectralgpt} models spatial-spectral dependencies in multispectral data via 3D generative pretraining.
However, most of these models are tailored for multispectral data, leaving a gap in hyperspectral applications. To address this,  \citep{maskedsst} introduced Masked Vision Transformers for Hyperspectral Image Classification. SpectralEarth \citep{spectralearth} stands out for integrating a spectral adapter into standard vision backbones, thereby allowing the model to better handle the distinct spectral characteristics of hyperspectral data. Additionally, HyperSIGMA  \citep{hypersigma} introduces sparse sampling attention (SSA), which effectively reduces spectral and spatial redundancies while enhancing the learning of diverse contextual features. This mechanism contributes to a versatile model capable of excelling across a wide range of hyperspectral image processing tasks, such as classification, change detection, unmixing, and anomaly detection.

\subsection{Parameter-Efficient Fine-Tuning (PEFT)}

Parameter-Efficient Fine-Tuning (PEFT) is a set of techniques designed to adapt large pre-trained models to new tasks while updating only a small fraction of the model's parameters. Formally, given a pre-trained model \(M\) parametrized by \(\theta\), and a downstream task \(D\), PEFT introduces a task-specific parameter increment \(\Delta \theta\) such that \(|\Delta \theta| \ll |\theta|\), allowing the model to be fine-tuned efficiently  \citep{peft_survey}. The goal is to minimize the task-specific loss by adjusting only the essential parameters, significantly reducing computational overhead and storage costs.
According to  \citep{peft_survey}, PEFT methods can be broadly classified into three categories: addition-based methods, unified-based tuning, and partial-based tuning. Addition-based methods introduce additional learnable modules into the model, as seen in techniques like Adapter Tuning \citep{adapter} and Visual Prompt Tuning \citep{vpt}. Unified-based tuning integrates multiple tuning methods into a single framework, such as NOAH \citep{noah}, which combines different adapters. Partial-based tuning involves updating only specific parts of the model’s parameters. This class includes Specification Tuning, where only certain parameters like biases are optimized (e.g., BitFit  \citep{bitfit}), and Reparameter Tuning, such as LoRA  \citep{lora}, which introduces low-rank matrices to reduce the number of trainable parameters. Another method, called KronA \citep{krona}, addresses the limitation of LoRA's low-rank constraint by utilizing a Kronecker product. This approach removes the low-rank restriction, allowing for greater expressiveness while maintaining efficiency. Lokr (Low-Rank Adaptation with Kronecker Product)  \citep{lycoris} is another method that combines KronA and LoRA, demonstrating significant capability in the efficient fine-tuning of Stable Diffusion models. 
Generally, PEFT methods like LoRA employ identical learning rates for adapter matrices A and B. The authors of LoRA+ \citep{lora+} investigated this and found that it can lead to suboptimal fine-tuning, especially for models with large embedding dimensions. To address this, they proposed LoRA+, which uses different learning rates for A and B, improving performance and increasing speed during fine-tuning.

\section{Methodology}

In this section, we detail the process of adapting the SpectralGPT foundation model for hyperspectral image classification and the parameter-efficient fine-tuning (PEFT) methods used to optimize the training process. 
\subsection{Overview of the pipeline}
\begin{figure*}[ht]
    \centering
    \includegraphics[width=\textwidth]{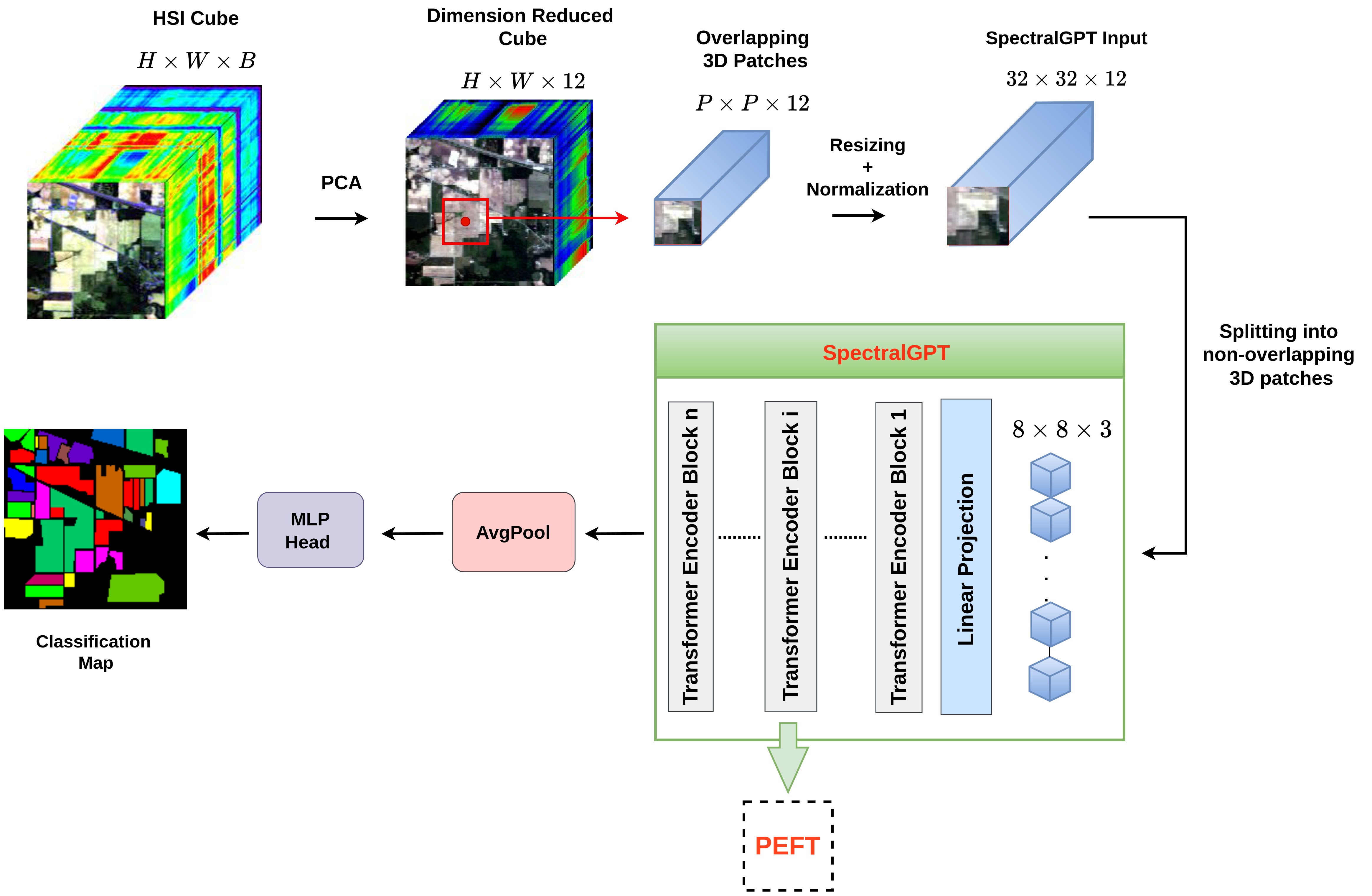}
    \scriptsize \caption{ \scriptsize An overview of the overall pipeline used to fine-tune the SpectralGPT model for hyperspectral image classification. It begins with the HSI Cube, which undergoes dimensionality reduction via PCA to produce a reduced cube of shape \( H \times W \times 12 \). This cube is divided into overlapping 3D patches, which are then resized and normalized. Each patch is subsequently split into overlapping \( 8 \times 8 \times 3 \) 3D tokens for input into SpectralGPT. SpectralGPT processes these tokens through transformer encoder blocks, followed by an average pooling layer (AvgPool) and an MLP head to produce the classification map. When we are not doing full finetuning, PEFT methods can be  applied at the level of each transformer encoder block}
\label{fig:hsic_pipeline}
\end{figure*}

The figure \textcolor{red}{\ref{fig:hsic_pipeline}} illustrates the global pipeline of our fine-tuning approach, highlighting the transformations applied from the raw HSI cube to the classification maps.

\begin{enumerate}
    \item[$\bullet$] Architecture of SpectralGPT: \\
    To achieve our objective of fine-tuning a pre-trained multispectral foundation model for hyperspectral image classification (HSIC), we utilized the SpectralGPT \citep{spectralgpt} model, which is specifically designed to handle spectral remote sensing data. 
    The model leverages the principles of Masked Autoencoder (MAE) for pretraining, where a spectral-wise 3D tensor masking is employed to mask 90\% of the 3D tokens. The encoder processes all the spatial-spectral visible tokens, while a lightweight decoder reconstructs the masked portions, allowing the model to learn robust and comprehensive representations from spectral data.
    
    At its core, the encoder of SpectralGPT is built on the vanilla ViT-Base architecture, with a token size of  $8 \times 8 \times 3$ to adapt the model to spectral data. SpectralGPT also employs two learnable positional embeddings: one dedicated to capturing spatial features and the other designed to learn patterns across spectral channels. 
    Additionally, SpectralGPT is highly flexible and capable of accommodating various input image sizes. It is progressively pre-trained on two significant multispectral datasets: the fMoW \citep{fmow} dataset with images of 
    $96 \times 96$ pixels, followed by the BigEarthNet \citep{bigearthnet} dataset, which contains images of  $128 \times 128$ pixels. This progressive training on varying resolutions makes SpectralGPT highly adaptable to different remote sensing data formats.
    To fine-tune for downstream tasks, an average pooling layer and a linear classification layer are added on top of the encoder.

     \item[$\bullet$] Band Reduction\\
    Since the SpectralGPT model is pre-trained on datasets with 12 spectral bands, we apply Principal Component Analysis (PCA) to reduce the number of spectral bands in each HSI dataset to 12. This ensures compatibility with the model while preserving most of the spectral variance in the data.

     \item[$\bullet$] Selection of Image Patch Size and Model Input Size\\
    The selection of the appropriate patch size is a well-known challenge in Hyperspectral Image Classification (HSIC), as it significantly affects model performance. Patch size determines the amount of local spatial and spectral context captured within each patch, directly impacting the ability of the model to leverage fine-grained information. In this work, we evaluated patch sizes of \(7 \times 7\), \(9 \times 9\), \(11 \times 11\), \(17 \times 17\), \(23 \times 23\), and \(27 \times 27\) for each dataset and selected the ones leading to the best performance.
    One advantage of the SpectralGPT model is its flexibility in accommodating different input sizes. In this study, all patches were resized to a uniform input size of \(32 \times 32\). The choice of \(32 \times 32\) is motivated by its ability to balance several key factors. Compatibility with architecture is a key consideration, as the SpectralGPT model, built on a Vision Transformer (ViT) backbone, tokenizes inputs into \(8 \times 8 \times 3\) (or \(16 \times 16 \times 3\)) patches. This necessitates an input size that is a multiple of 16 and 8, ensuring seamless integration with the model's architecture. Preservation of spatial-spectral context is also critical in HSIC. Resizing introduces a trade-off between maintaining spatial detail and avoiding spectral distortion. Smaller resized inputs overly compress spatial features, resulting in a reduction of critical spatial-spectral information necessary for effective classification. In contrast, larger resized inputs risk introducing interpolation artifacts that can distort spectral fidelity or introduce noise. The  $32 \times 32 $ size offers a balanced representation, preserving both spatial and spectral information while minimizing distortion during resizing.
    Moreover, given the constraints of our single GPU setup, increasing the input size—such as to \(64 \times 64\)—resulted in insufficient memory issues. In this context, the \(32 \times 32\) input size strikes a balance between preserving spatial-spectral richness and meeting memory requirements, enabling efficient training without running into memory limitations. This consistent resizing ensures a streamlined pipeline across datasets, minimizing architectural changes and facilitating comparative analysis.
    
     \item[$\bullet$] Normalization Strategies\\
    In terms of normalization, we initially used standardization with the mean and standard deviation specific to each of the 12 bands in BigEarthNet dataset. When this method did not lead to satisfactory performance, we switched to min-max normalization, which provided better results for some datasets.
    
\end{enumerate}

\subsection{PEFT Methods}
\begin{figure*}[ht]
    \centering
    \includegraphics[width=\textwidth]{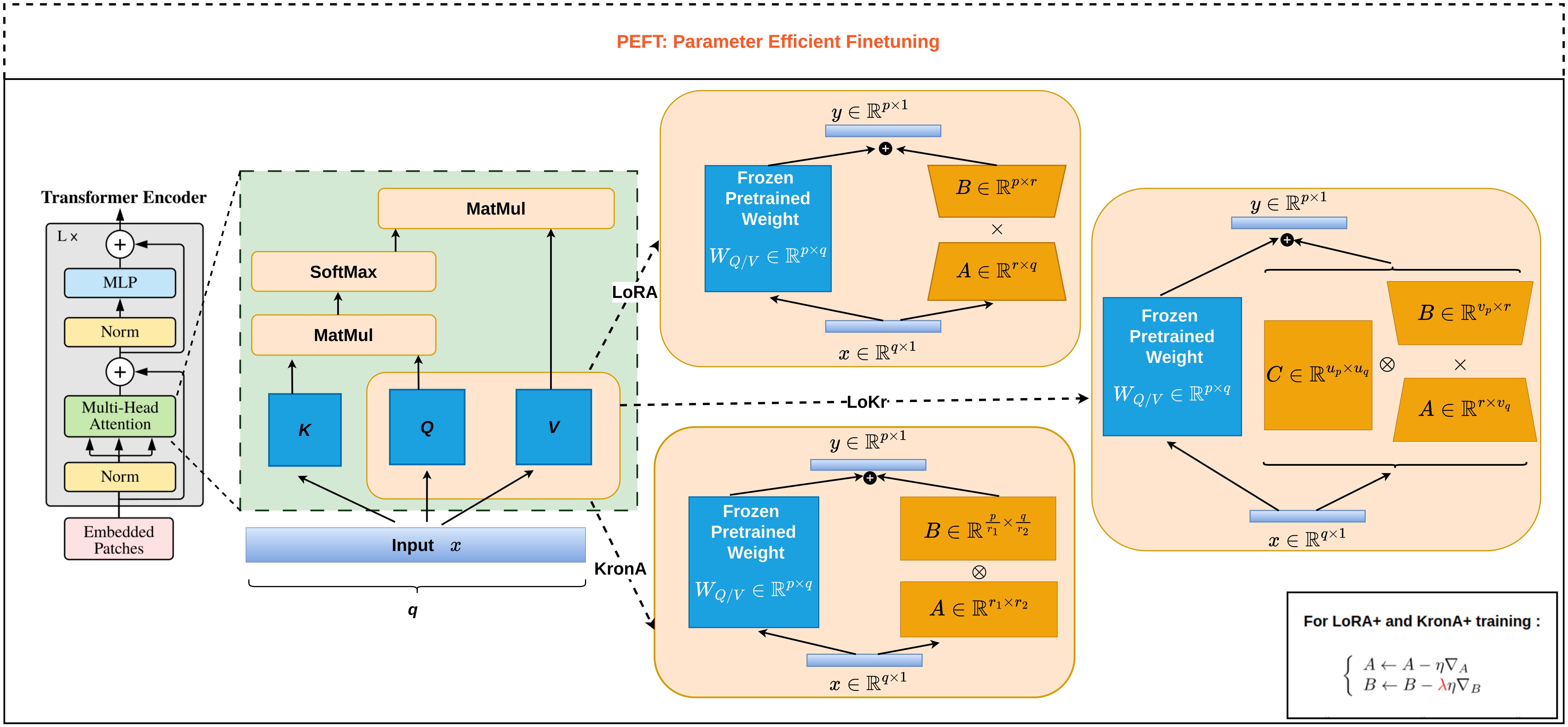}
     \caption{ \scriptsize The Parameter-Efficient Fine-Tuning (PEFT) strategies used in the study: LoRA, KronA, and LoKr are applied to the \( Q \) and \( V \) components of each transformer layer to adapt the frozen pre-trained weights with minimal additional parameters. The rectangle at the bottom-right highlights the parameter update strategy used in the methods of LoRA+ and KronA+, where instead of using the same learning rate for \( A \) and \( B \) as in LoRA and KronA, the learning rate of \( B \) is set to be \( \lambda \times \) that of \( A \), with \( \lambda \gg 1 \) fixed.}

\label{fig:peft_structure}
\end{figure*}

As illustrated in the figure \ref{fig:peft_structure} We apply parameter-efficient fine-tuning (PEFT) methods only to the weights of the query (Q) and value (V) components of all the multi-head self-attention (MHSA) layers within the encoder. The following subsections describe the different PEFT methods applied in this process.

\subsubsection{LoRA}
        Low-Rank Adaptation (LoRA) reduces the number of trainable parameters by introducing low-rank updates to the weight matrices of the pre-trained model. The key idea is to decompose the weight matrix \( W \) into two smaller low-rank matrices \( A \in R^{r \times q} \) and \( B \in R^{p \times r} \). The weight update \(\Delta W\) is computed as the product of the two low-rank matrices:
        \begin{equation}
         \Delta W = BA 
        \end{equation}
        
        A scaling factor \(\alpha\) is introduced to  control the contribution of the low-rank updates to
        the original weight matrix, modifying the weight update to:
        \begin{equation}
            \Delta W = \frac{\alpha}{r}BA
        \end{equation}
        
        The output \(y\) is computed by adding the weight update \(\Delta W\) to the original weight matrix \( W_0 \) and multiplying the result by the input \( x \):
        \begin{equation}
            y = W_0 x + \Delta W x = W_0 x + \frac{\alpha}{r} B A x
        \end{equation}

    \subsubsection{KronA}
    
        KronA (Kronecker Adapter) \citep{krona} improves upon LoRA by addressing the limitations of low-rank factorization, which can sometimes lack sufficient representation power. KronA replaces the low-rank approximation with a \textbf{Kronecker product}, which decomposes the weight matrix into two smaller matrices.
        Given matrices \( A \in R^{r_1 \times r_2} \) and \( B \in R^{\frac{p}{r_1} \times \frac{p}{r_2} }  \), the \textbf{Kronecker product} \citep{kronproduct} is defined as:
        
        \begin{equation}
            A \otimes B = \begin{pmatrix}
        a_{11} B & \dots & a_{1r_2} B \\
        \vdots & \ddots & \vdots \\
        a_{r_11} B & \dots & a_{r_1r_2} B
        \end{pmatrix}
        \end{equation}
        
        The hyperparameters \( r_1 \) and \( r_2 \) control the number of trainable parameters and the rank of \( \Delta W \). 
        
        The weight update \( \Delta W \) is defined as:
        
        \begin{equation}
            \Delta W = s \cdot (A \otimes B)
        \end{equation}
        
        where \( s \) is a scaling factor that adjusts the contribution of the Kronecker product to the weight matrix update.
        The final output \( y \) is computed by incorporating the weight update \( \Delta W \) and multiplying by the input \( x \):
        
        \begin{equation}
             y = W_0 x + \Delta W x = W_0 x + s \cdot (A \otimes B) x
        \end{equation}

     \subsubsection{LoKr}
        Low-Rank Adaptation with Kronecker Product (LoKr) is an extension of the KronA method, leveraging the Kronecker product to efficiently decompose weight matrices and enhance parameter efficiency. By combining the rank-multiplicative nature of the Kronecker product with low-rank decomposition, LoKr provides greater flexibility and expressiveness for fine-tuning large models.
        
        LoKr introduces a user-defined hyperparameter, the factor \( f \), which determines the dimensions of the sub-weights used in the Kronecker decomposition. Specifically, \( f \) defines the sizes of the sub-weights as follows:
        \begin{equation}
          \begin{cases}
              u_p = \max( u \leq \min(f, \sqrt{p}) | p \mod u = 0), \quad v_p = \frac{p}{u_p}, \\
            u_q = \max( u \leq \min(f, \sqrt{q}) | q \mod u = 0), \quad v_q = \frac{q}{u_q}
          \end{cases}
        \end{equation}
        
        This relationship ensures that \( u_p \) and \( u_q \) are the largest valid divisors of \( p \) and \( q \), respectively, that do not exceed the smaller of \( f \) and \( \sqrt{p} \) (or \( \sqrt{q} \)). This helps optimize the rank and size of the sub-weights while ensuring that the right block is the larger of the two. Using these dimensions, the two sub-weights \( C \in R^{u_p \times u_q} \) and \( C_1 \in R^{v_p \times v_q} \) are constructed, and the weight update \( \Delta W \) is calculated as:
        \begin{equation}
            \Delta W = \gamma \; (C \otimes C_1)
        \end{equation}
    
        where \( \gamma \) is a scaling coefficient that controls the contribution of the Kronecker-based update to the original weight matrix.
        A second user-defined hyperparameter, \( r \), further incorporates a low-rank decomposition into the right block \( C_1 \), refining it into two smaller matrices \( A \in R^{r \times v_q} \) and \( B \in R^{v_p \times r} \). The updated weight matrix becomes:
        \begin{equation}
            \Delta W = \gamma \; (C \otimes (B A))
        \end{equation}
        The final output \( y \) is computed by adding the weight update \( \Delta W \) to the initial weight matrix \( W_0 \):
        \begin{equation}
            y = W_0 x + \Delta W x = W_0 x + \gamma \; (C \otimes (B A)) x
        \end{equation}

    \subsubsection{LoRA+ and KronA+}

        LoRA+ builds upon the original Low-Rank Adaptation (LoRA) method by modifying the learning rate scheme for the low-rank matrices, leading to more effective fine-tuning. In the standard LoRA, two low-rank matrices \( A \) and \( B \) are introduced to reduce the number of trainable parameters. During training, the updates to these matrices are performed using the same learning rate \( \eta \), which can lead to suboptimal learning, especially in models with large embedding dimensions. Mathematically, the parameter updates in standard LoRA are defined as:

        \begin{equation}
		\begin{cases}
			A \leftarrow A - \eta \nabla_A \\
			B \leftarrow B - \eta \nabla_B
		\end{cases}
	\end{equation}
        
        where \( \nabla_A \) and \( \nabla_B \) represent the gradients of \( A \) and \( B \), respectively, calculated using an optimizer like AdamW.
        
        LoRA+ addresses this by applying a scaled learning rate to matrix \( B \). Specifically, the learning rate for \( B \) is set to \( \lambda \eta \), where \( \lambda > 1 \) is a hyperparameter that amplifies the learning rate for \( B \). This approach enhances the model's performance while preserving efficient parameter updates. The update equations for LoRA+ are:
        \begin{equation}
            \begin{cases}
                A \leftarrow A - \eta \nabla_A \\
                B \leftarrow B - \textcolor{red}{\lambda} \eta \nabla_B
            \end{cases}
        \end{equation}
        
        where \( \textcolor{red}{\lambda} \) is a fixed multiplier that adjusts the learning dynamics between \( A \) and \( B \).
        
        We apply the same principle to develop KronA+, an extension of KronA that uses different learning rates for its Kronecker product-based low-rank matrices, enhancing the model's performance while maintaining efficient training.

    \subsubsection{Computational Complexity}

        The computational complexity of each fine-tuning method is determined by the way it updates the \( Q \) (query) and \( V \) (value) attention components. Given a transformer with \( L \) attention layers and hidden size \( d \), Full Fine-Tuning updates all parameters, leading to a complexity of \( \mathcal{O}(L d^2) \). BitFit, which modifies only biases, reduces this to \( \mathcal{O}(L d) \). LoRA introduces low-rank decomposition with complexity \( \mathcal{O}(L r d) \), where \( r \) is the decomposition rank. KronA applies Kronecker factorization, resulting in \( \mathcal{O}(L (r_1 r_2 + \frac{p}{r_1} \cdot \frac{p}{r_2})) \), where \( r_1, r_2, p \) define Kronecker sub-weights. LoKr further refines this structure with \( \mathcal{O}(L (u_p u_q + r (v_p + v_q))) \), where \( u_p, u_q, v_p, v_q \) represent the dimensions of its factorized components.

\section{Experiment Setup}

\subsection{Datasets}

To ensure a comprehensive evaluation of our proposed approach, we use a diverse set of hyperspectral image (HSI) datasets captured by different sensors. These datasets are widely recognized in the hyperspectral classification domain and have been extensively used as benchmarks in previous studies. The selected datasets include Indian Pines, Pavia University, Houston 2013, Botswana, and WHU-Hi Longkou.

In the following subsections, we describe each dataset and present its training and testing pixel distributions to facilitate reproducibility in future research. To ensure a fair evaluation, training and testing pixels are selected disjointly, with the test set being larger than the training set. This setup provides a better assessment of the model’s generalization ability, as it must learn from limited labeled samples while being evaluated on a broader distribution. Moreover, this approach aligns with real-world hyperspectral imaging scenarios, where labeled data is scarce, but large-scale unlabeled imagery is readily available for analysis.

\begin{figure}[ht]
    \centering
        \centering
        \includegraphics[height=5cm, keepaspectratio]{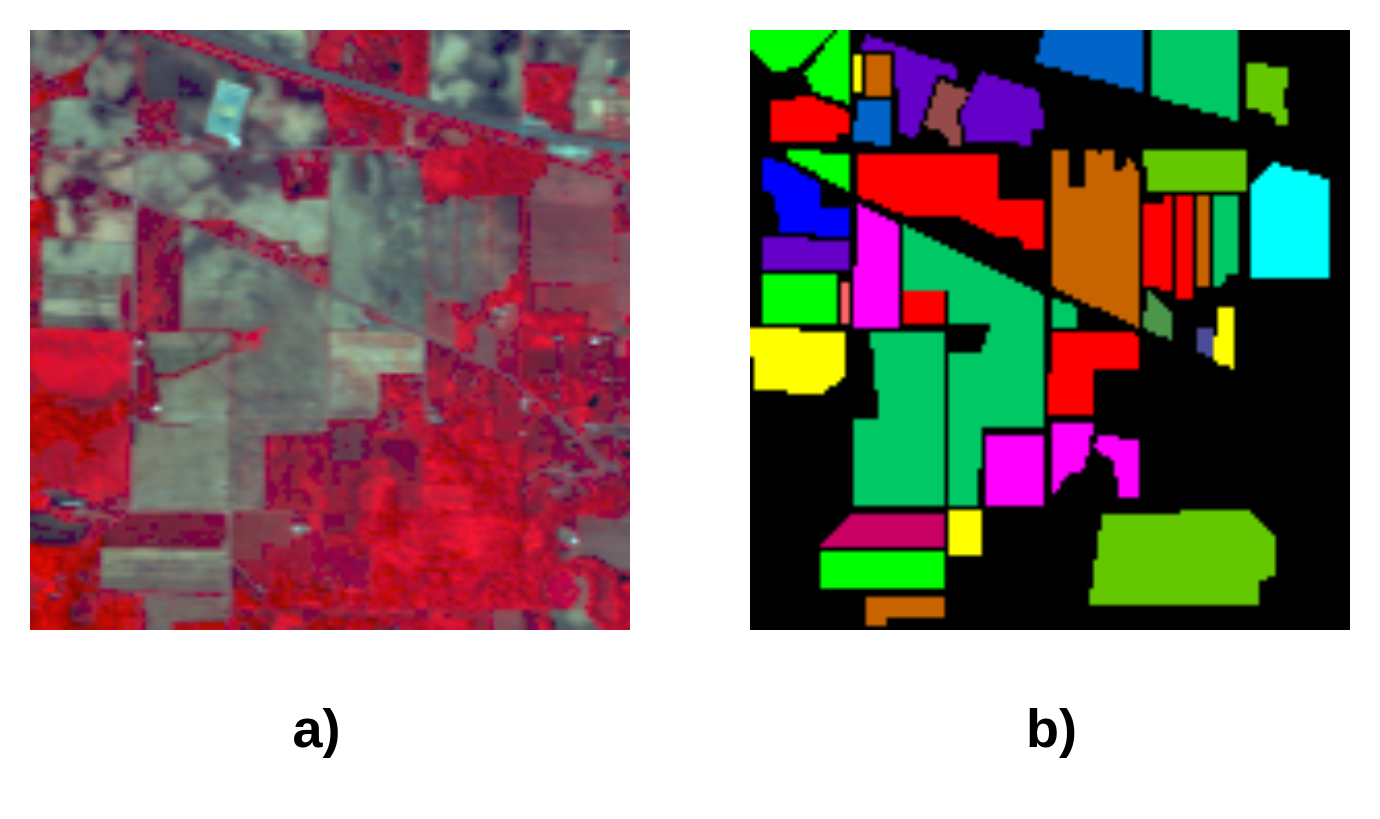} 
    \caption{\scriptsize Visualization of the Indian Pines dataset. (a) False color Image. (b) Ground Truth (GT)}
    \label{fig:FC_GT_IP}
\end{figure}

\begin{figure}[ht]
    \centering
    \includegraphics[height=6cm, keepaspectratio]{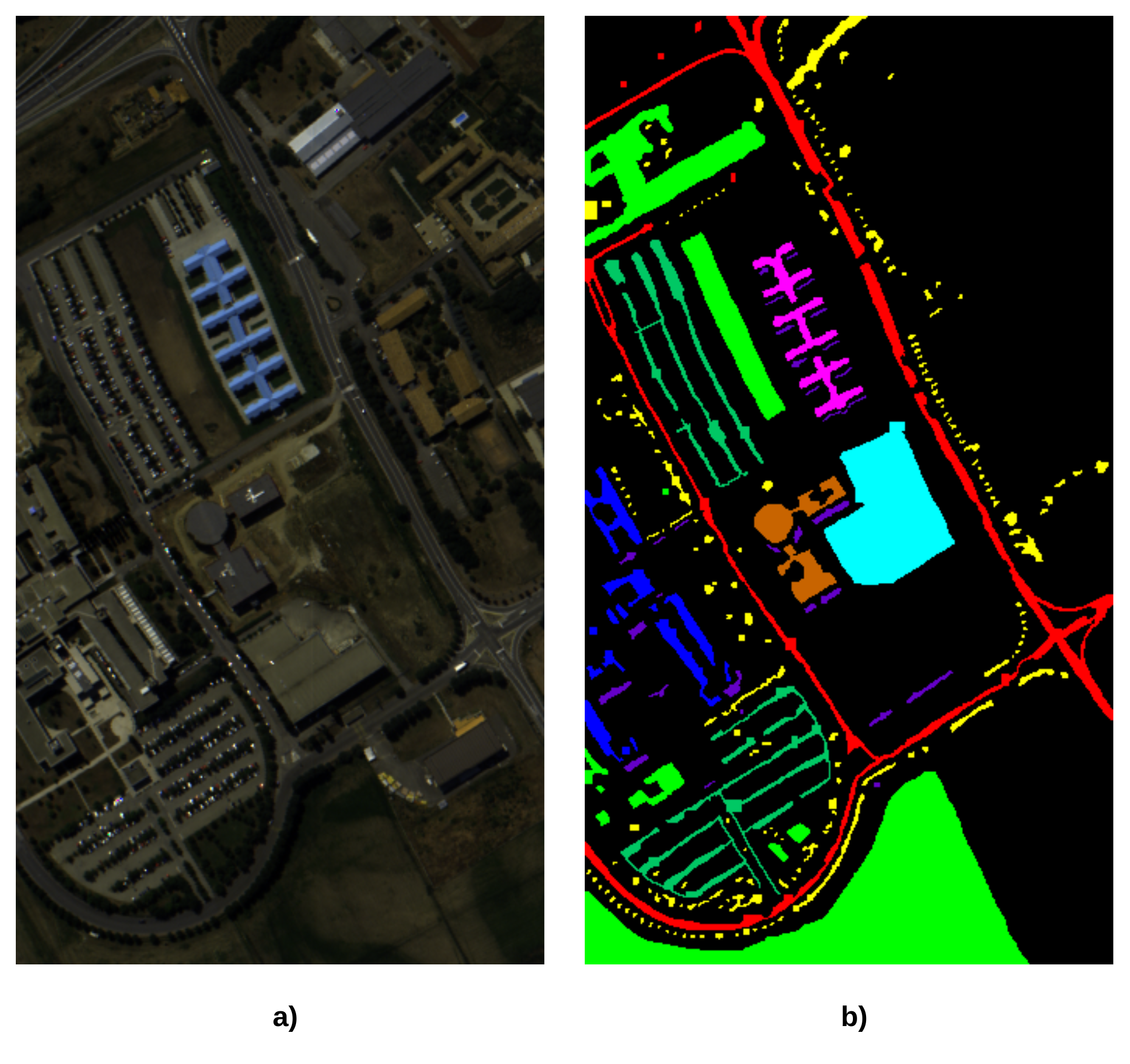} 
    \caption{\scriptsize Visualization of the Pavia dataset. (a) False color Image. (b) Ground Truth (GT)}
    \label{fig:FC_GT_pavia}
\end{figure}

\begin{figure}[ht]
    \centering
        \includegraphics[height=6cm, keepaspectratio]{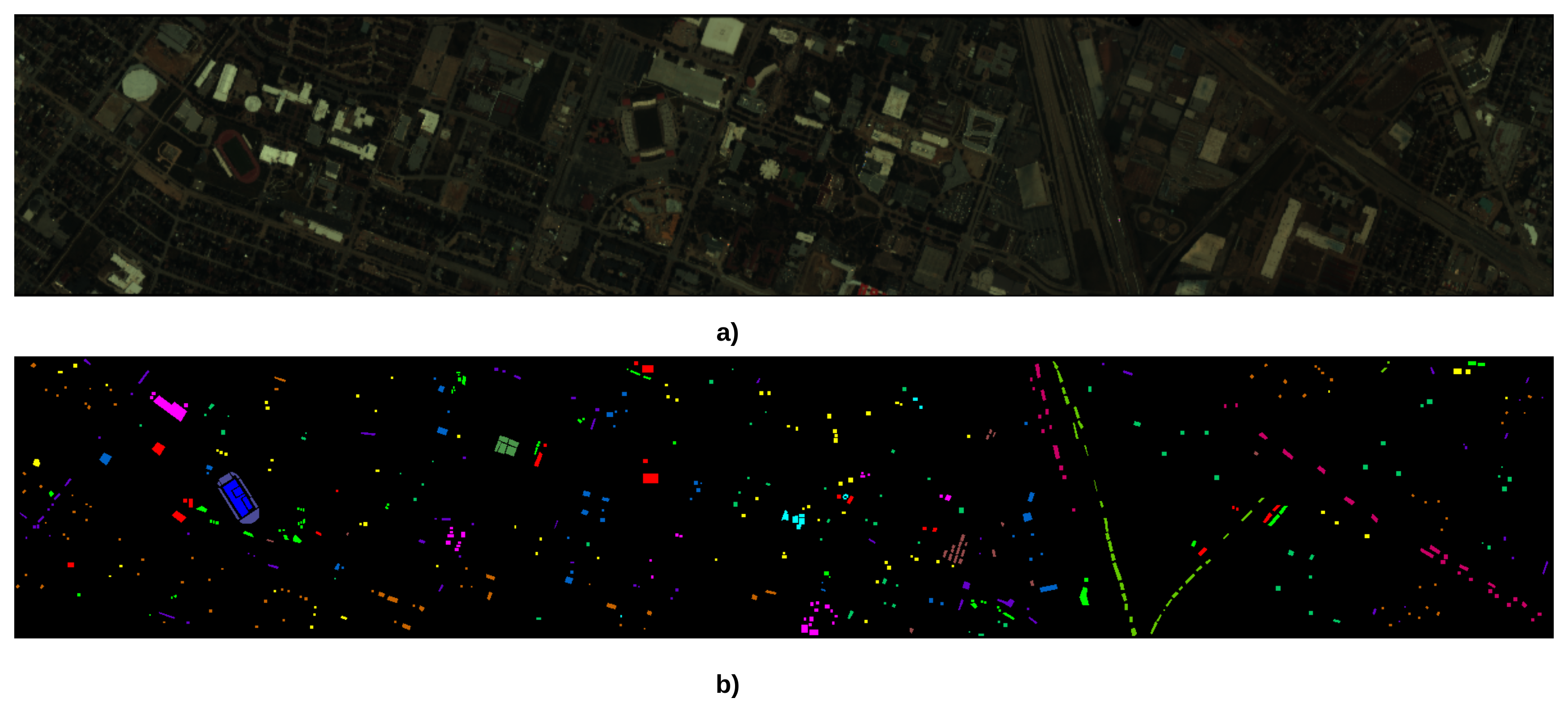} 
    \caption{\scriptsize Visualization of the Houston dataset. (a) False color Image. (b) Ground Truth (GT)}
    \label{fig:FC_GT_houston}
\end{figure}

\begin{figure}[ht]
    \centering
        \includegraphics[height=9cm, keepaspectratio]{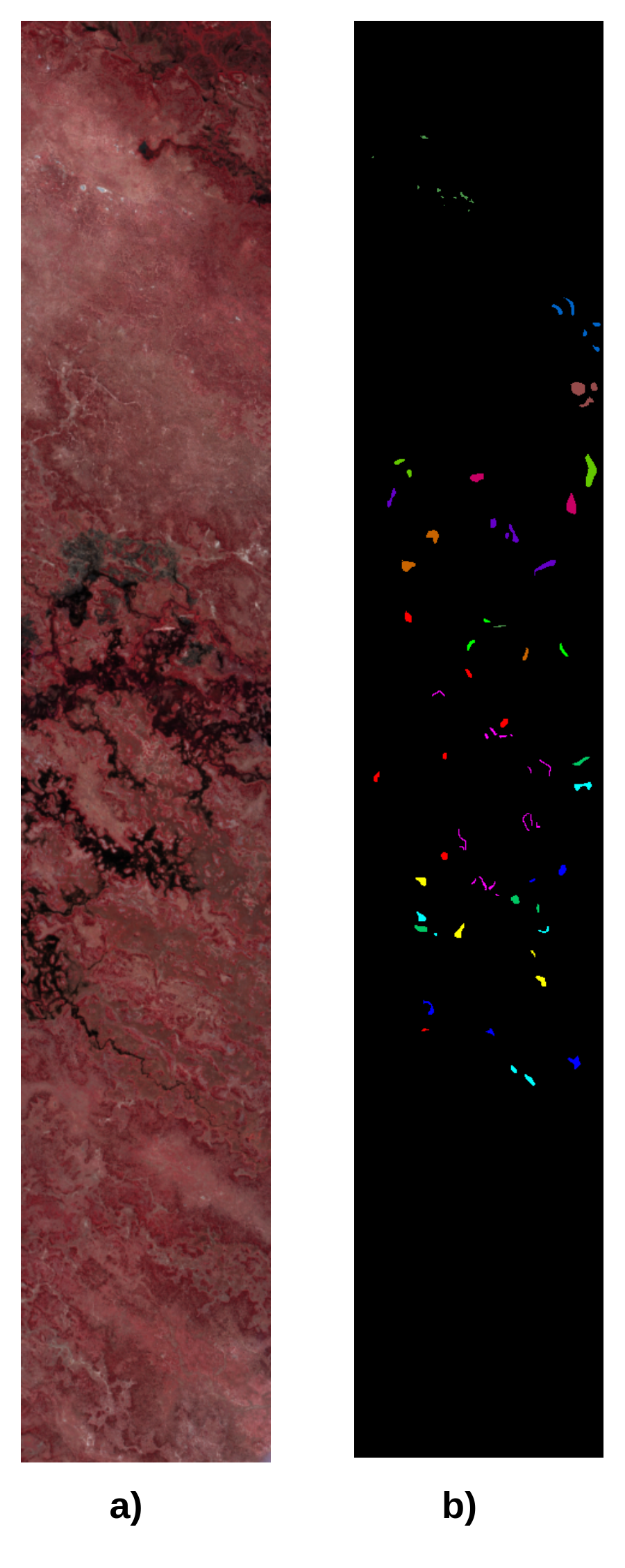} 
    \caption{\scriptsize Visualization of the Botswana dataset. (a) False color Image. (b) Ground Truth (GT)}
    \label{fig:FC_GT_Botswana}
\end{figure}

\begin{figure}[ht]
    \centering
        \includegraphics[height=5cm, keepaspectratio]{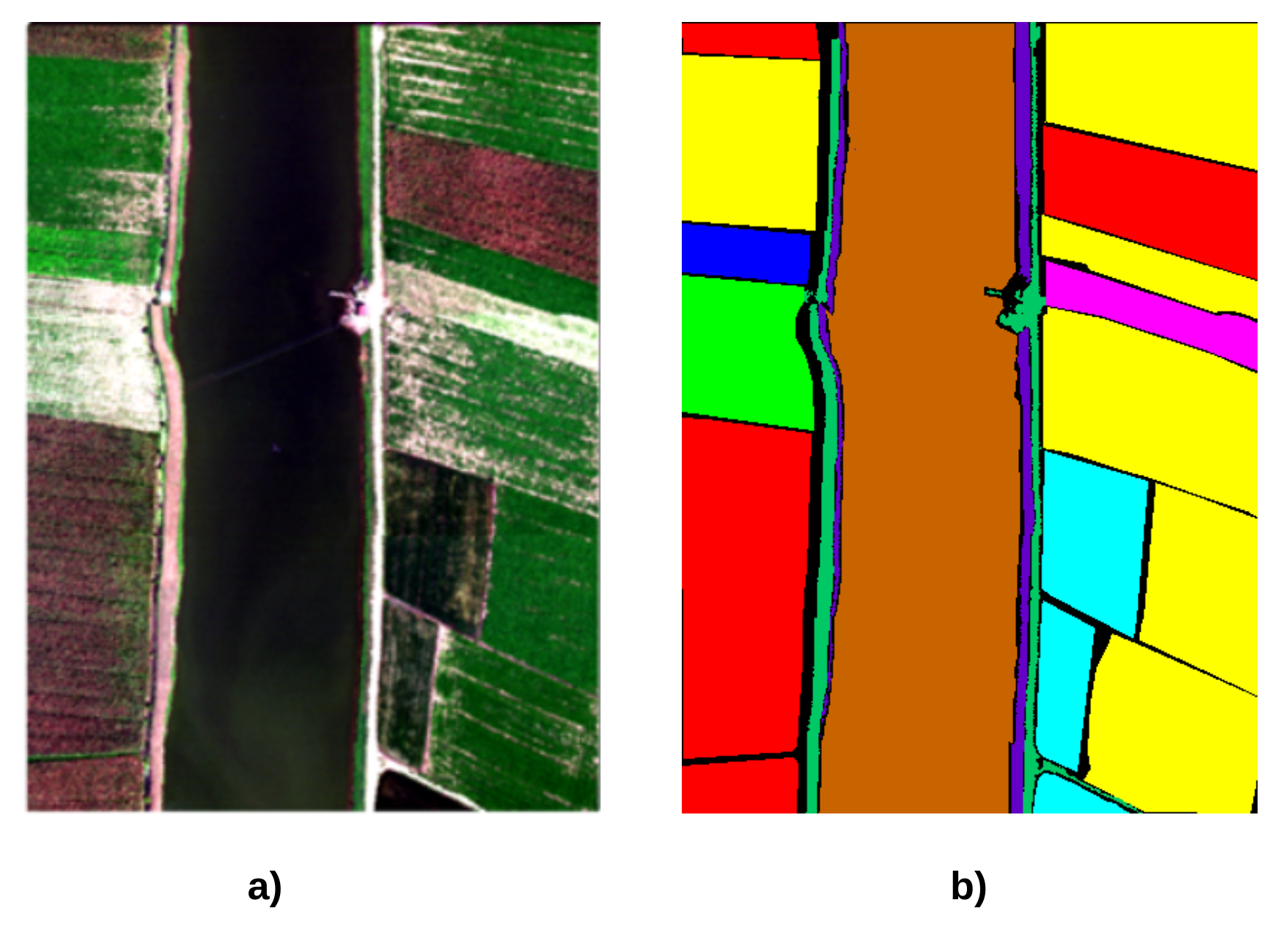} 
    \caption{\scriptsize Visualization of the Longkou dataset. (a) False color Image. (b) Ground Truth (GT)}
    \label{fig:FC_GT_longkou}
\end{figure}

Figures \ref{fig:FC_GT_IP}, \ref{fig:FC_GT_pavia},\ref{fig:FC_GT_houston}, \ref{fig:FC_GT_Botswana} and  \ref{fig:FC_GT_longkou} show in (a) the False Color image and in (b) the Ground-Truth for each dataset. Additionally, Tables \ref{tab:indian_pines_split}, \ref{tab:pavia_split}, \ref{tab:houston_split}, \ref{tab:botswana_split} and \ref{tab:longkou_split} present the class-wise distribution of pixels in the training and testing sets, ensuring transparency and reproducibility of the experimental setup.

\subsubsection{Indian Pines}
Captured by the AVIRIS(Airborne Visible/Infrared Imaging Spectrometer)  \citep{aviris} sensor over agricultural fields in northwestern Indiana, this dataset comprises 145x145 pixels with 224 spectral bands ranging from 0.4 to 2.5$\mu$m. It includes 16 land cover classes, primarily featuring crops like corn and soybeans in early growth stages. 

\begin{table}[ht]
    \centering
    \small 
    \setlength{\tabcolsep}{4pt} 
    \caption{\scriptsize Class-wise distribution of pixels in the Indian Pines training and testing sets.}
    \label{tab:indian_pines_split}
    \begin{tabular}{lccc}
        \toprule
        \textbf{Class No.} & \textbf{Class Name} & \textbf{Training Pixels} & \textbf{Testing Pixels} \\
        \midrule
        1  & Alfalfa                             & 50  & 1384 \\
        2  & Corn-notill                         & 50  & 784  \\
        3  & Corn-mintill                        & 50  & 184  \\
        4  & Corn                                & 50  & 447  \\
        5  & Grass-pasture                       & 50  & 697  \\
        6  & Grass-trees                         & 50  & 439  \\
        7  & Grass-pasture-mowed                 & 50  & 918  \\
        8  & Hay-windrowed                       & 50  & 2418 \\
        9  & Oats                                & 50  & 564  \\
        10 & Soybean-notill                      & 50  & 162  \\
        11 & Soybean-mintill                     & 50  & 1244 \\
        12 & Soybean-clean                       & 50  & 330  \\
        13 & Wheat                               & 50  & 45   \\
        14 & Woods                               & 15  & 39   \\
        15 & Buildings-Grass-Trees-Drives        & 15  & 11   \\
        16 & Stone-Steel-Towers                  & 15  & 5    \\
        \midrule
        \textbf{Total} &  & \textbf{695} & \textbf{9671} \\
        \bottomrule
    \end{tabular}
\end{table}

\subsubsection{Pavia University}
This ROSIS (Reflective Optics System Imaging Spectrometer)  \citep{rosis} dataset was captured during a flight campaign over Pavia, Nothern Italy. The image covers 610×340 pixels with 115 spectral bands and 9 classes. 

\begin{table}[ht]
    \centering
    \small 
    \setlength{\tabcolsep}{4pt} 
    \caption{\scriptsize Class-wise distribution of pixels in the Pavia training and testing sets.}
    \label{tab:pavia_split}
    \begin{tabular}{lccc}
        \toprule
        \textbf{Class No.} & \textbf{Class Name} & \textbf{Training Pixels} & \textbf{Testing Pixels} \\
        \midrule
        1  & Asphalt                & 548  & 6304  \\
        2  & Meadows                & 540  & 18146 \\
        3  & Gravel                 & 392  & 1815  \\
        4  & Trees                  & 524  & 2912  \\
        5  & Painted Metal Sheets   & 265  & 1113  \\
        6  & Bare Soil              & 532  & 4572  \\
        7  & Bitumen                & 375  & 981   \\
        8  & Self-Blocking Bricks   & 514  & 3364  \\
        9  & Shadows                & 231  & 795   \\
        \midrule
        \textbf{Total} &  & \textbf{3921} & \textbf{40002} \\
        \bottomrule
    \end{tabular}
\end{table}

\subsubsection{Houston2013 Dataset}
    The Houston2013 dataset is a widely-used hyperspectral dataset collected by the ITRES CASI-1500 sensor over the University of Houston campus and its neighboring urban areas in Texas, USA. The dataset consists of 144 spectral bands spanning the 0.364 to 1.046$\mu$m wavelength range, with 10 nm intervals. It has a spatial resolution of 2.5 meters, and the hyperspectral cube contains $349 \times 1905$ pixels. The dataset includes 15 urban land-cover classes, such as asphalt, grass, trees, and residential buildings, making it a challenging benchmark for land-cover classification.

\begin{table}[ht]
    \centering
    \small 
    \setlength{\tabcolsep}{4pt} 
    \caption{\scriptsize Class-wise distribution of pixels in the Houston training and testing sets.}
    \label{tab:houston_split}
    \begin{tabular}{lccc}
        \toprule
        \textbf{Class No.} & \textbf{Class Name} & \textbf{Training Pixels} & \textbf{Testing Pixels} \\
        \midrule
        1  & Healthy Grass       & 198  & 1053  \\
        2  & Stressed Grass      & 190  & 1064  \\
        3  & Synthetic Grass     & 192  & 505   \\
        4  & Tree                & 188  & 1056  \\
        5  & Soil                & 186  & 1056  \\
        6  & Water               & 182  & 143   \\
        7  & Residential         & 196  & 1072  \\
        8  & Commercial          & 191  & 1053  \\
        9  & Road                & 193  & 1059  \\
        10 & Highway             & 191  & 1036  \\
        11 & Railway             & 181  & 1054  \\
        12 & Parking Lot1        & 192  & 1041  \\
        13 & Parking Lot2        & 184  & 285   \\
        14 & Tennis Court        & 181  & 247   \\
        15 & Running Track       & 187  & 473   \\
        \midrule
        \textbf{Total} &  & \textbf{2832} & \textbf{12197} \\
        \bottomrule
    \end{tabular}
\end{table}

 \subsubsection{Botswana}
 This dataset was captured by the Hyperion sensor aboard NASA's EO-1 satellite over the Okavango Delta, Botswana. It consists of 242 spectral bands in the 0.4 to 2.5$\mu$m range, covering $1476 \times 256$ pixels areas with 14 ground-truth classes.

\begin{table}[ht]
    \centering
    \small 
    \setlength{\tabcolsep}{4pt} 
    \caption{\scriptsize Class-wise distribution of pixels in the Botswana training and testing sets.}
    \label{tab:botswana_split}
    \begin{tabular}{lccc}
        \toprule
        \textbf{Class No.} & \textbf{Class Name} & \textbf{Training Pixels} & \textbf{Testing Pixels} \\
        \midrule
        1  & Water                  & 30  & 240  \\
        2  & Hippo grass            & 30  & 71   \\
        3  & Floodplain grasses 1   & 30  & 221  \\
        4  & Floodplain grasses 2   & 30  & 185  \\
        5  & Reeds                  & 30  & 239  \\
        6  & Riparian               & 30  & 239  \\
        7  & Firescar               & 30  & 229  \\
        8  & Island interior        & 30  & 173  \\
        9  & Acacia woodlands       & 30  & 284  \\
        10 & Acacia shrublands      & 30  & 218  \\
        11 & Acacia grasslands      & 30  & 275  \\
        12 & Short mopane           & 30  & 151  \\
        13 & Mixed mopane           & 30  & 238  \\
        14 & Exposed soils          & 30  & 65   \\
        \midrule
        \textbf{Total} &  & \textbf{420} & \textbf{2828} \\
        \bottomrule
    \end{tabular}
\end{table}

\subsubsection{WHU-Hi-LongKou dataset} 
 This dataset was acquired over Longkou Town, Hubei province, using the same UAV platform equipped with a Headwall Nano-Hyperspec sensor. The surveyed area consists of simple agricultural fields with crops such as corn, cotton, sesame, and different types of soybean. The images were collected at an altitude of 500 m, yielding data with 270 spectral bands between 0.4 to 2.5$\mu$m, a spatial resolution of 0.463 m, and an image size of 550 $\times$ 400 pixels.

\begin{table}[ht]
    \centering
    \small 
    \setlength{\tabcolsep}{4pt} 
    \caption{\scriptsize Class-wise distribution of pixels in the Longkou training and testing sets.}
    \label{tab:longkou_split}
    \begin{tabular}{lccc}
        \toprule
        \textbf{Class No.} & \textbf{Class Name} & \textbf{Training Pixels} & \textbf{Testing Pixels} \\
        \midrule
        1  & Corn                  & 1000  & 33511  \\
        2  & Cotton                & 1000  & 7374   \\
        3  & Sesame                & 1000  & 2031   \\
        4  & Broad-leaf soybean    & 1000  & 62212  \\
        5  & Narrow-leaf soybean   & 1000  & 3151   \\
        6  & Rice                  & 1000  & 10854  \\
        7  & Water                 & 1000  & 66056  \\
        8  & Roads and houses      & 1000  & 6124   \\
        9  & Mixed weed            & 1000  & 4229   \\
        \midrule
        \textbf{Total} &  & \textbf{9000} & \textbf{195542} \\
        \bottomrule
    \end{tabular}
\end{table}

\newpage
\subsection{\textbf{Comparison with Baselines and other State-of-the-art( SOTA) Backbone Networks}}
In this study, before implementing full fine-tuning, LoRA, KronA, and LoKr adaptations, we set up Linear Probe and BitFit as baseline methods. 

\begin{enumerate}
    \item[$\bullet$] In \textbf{Linear Probe}, all parameters of the pre-trained model are frozen, with only a newly introduced linear classification head being trained.
    \item[$\bullet$] \textbf{BitFit} \citep{bitfit} is a specialized fine-tuning approach where only the model’s bias terms are fine-tuned, keeping all other parameters frozen. Specifically, we update only the bias terms in the query (\(Q\)) and value (\(V\)) components of each attention head.
\end{enumerate} 

Moreover, we compare our method against three prominent backbone networks specifically designed for hyperspectral image (HSI) classification: SpectralFormer, MambaHSI, and HyperSIGMA. To ensure a fair comparison, we strive to match the optimal hyperparameters used in the official implementations of these architectures as closely as possible.
\begin{enumerate}
    \item[$\bullet$] \textbf{SpectralFormer}  \citep{spectralformer}: SpectralFormer is a transformer-based network tailored for hyperspectral image classification. It introduces two key modules: Group-wise Spectral Embedding (GSE) to capture local spectral detail representation and Cross-layer Adaptive Fusion (CAF), a middle-range skip connection mechanism designed to adaptively fuse features across layers. In this study, SpectralFormer is trained with a batch size of 64, a patch size of 7, and a learning rate of \(5 \times 10^{-4}\). The number of training epochs varies across datasets: 300 for Botswana and Indian Pines, 600 for Houston and Pavia, and 100 for Longkou.

    \item[$\bullet$] \textbf{MambaHSI}  \citep{mambahsi}: MambaHSI leverages a state-space model (SSM) structure to capture long-range dependencies in hyperspectral data. Its end-to-end framework comprehensively incorporates spatial and spectral information. For all datasets in this study, MambaHSI is trained using a learning rate of \(3 \times 10^{-4}\) and 200 epochs.

    \item[$\bullet$] \textbf{HyperSIGMA}  \citep{hypersigma}: HyperSIGMA is a foundation model built specifically for hyperspectral imagery. It employs a sparse sampling attention mechanism to learn diverse contextual features and integrates a spectral enhancement module for spatial-spectral feature fusion. In this study, HyperSIGMA is trained across all datasets with a batch size of 64, a learning rate of \(6 \times 10^{-5}\), 200 epochs, and a patch size of 33. After dimensionality reduction, the number of bands is fixed to 30.
\end{enumerate}

\subsection{\textbf{Implementation Details}} 
The patch sizes for the Indian Pines, Pavia, Houston, Botswana and Longkou datasets are set to 15, 9, 9, 9 and 27 respectively, based on a detailed study of the effect of different patch sizes on the performance of each dataset. Data augmentation techniques, including horizontal and vertical flipping, radiation noise, and mixture noise, are used to enhance model robustness. For normalization, standardization using the mean and standard deviation of the BigEarthNet dataset is applied to the Indian Pines, Houston, Pavia, and Botswana datasets, while min-max normalization is used for the Longkou dataset.

Our models are implemented using the PyTorch library and trained on a Linux-based workstation equipped with an AMD Ryzen 9 7950X 16-Core Processor, 32GB of RAM, and an NVIDIA GeForce RTX 4080 SUPER GPU. All models are trained for 50 epochs using the AdamW \citep{adamw} optimizer with a weight decay of 0.05 and a batch size of 64. For full fine-tuning and linear probe methods, a learning rate of \(5 \times 10^{-5}\) is used, while a learning rate of \(5 \times 10^{-3}\) is applied for BitFit, LoRA, KronA, and LoKr.

For LoRA, we use a rank \(r = 4\) with a scaling factor \(\alpha = 4\), resulting in \(\alpha / r = 1\). Similarly, for KronA and LoKr, the parameters controlling the contribution of PEFT to the original weight (\(s\) for KronA and \(\gamma\) for LoKr) are set to 1, as the effect of these parameters is not studied in this work. 
For KronA, the shape of the matrice A (after tuning) is set to (384,2) for Indian Pines, Botswana and Longkou datasets, and to (24,32) and (16,48) respectively for Pavia and Houston datasets. The shape of the corresponding B is in the reverse order of A in each case. In LoKr, the factor \(f\) is set to 8, and a rank \(r = 8\) is applied for the low-rank decomposition of the right block resulting from the Kronecker decomposition. For the initialization of PEFT factors, one of the matrices is set as a zero matrix, while the other is initialized with values drawn from a Gaussian distribution \(N(0, \sigma^2)\).

\section{Results and Discussion}

\subsection{Quantitative evaluation}
The models are evaluated using three commonly used performance metrics in HSIC: Overall Accuracy (OA), Average Accuracy (AA), and the Kappa coefficient.
The quantitative classification results (OA, AA, kappa) and the number of parameters are presented in Tables \ref{table:IP}, \ref{table:pavia}, \ref{table:houston}, \ref{table:botswana}, \ref{table:longkou} 
for the Indian Pines, Pavia University, Houston, Botswana, and Longkou datasets, respectively.The best OA, AA, and Kappa are highlighted in \textcolor{red}{red} for PEFT methods and in \textcolor{blue}{blue} for the group of three SOTA backbones. The results of FFT SpectralGPT are shown in \textbf{bold}.

\subsubsection{Comparison with SOTA Backbones}  
The MambaHSI model highlights the strength of state-space models for HSI classification, outperforming Spectralformer and Hypersigma on all datasets except Botswana, where the latter two perform slightly better. Moreover, MambaHSI achieves higher performance than our FFT SpectralGPT on Indian Pines and WHU-Hi-LongKou, with differences of +10.15\% and +2.07\% in overall accuracy (OA), respectively. However, FFT SpectralGPT surpasses MambaHSI in the remaining three datasets, achieving, for instance, a +3.98\% OA advantage on the Houston dataset. This FFT SpectralGPT also systematically outperforms the Spectralformer model on all datasets except WHU-Hi-LongKou demonstrating its competitive spatial-spectral modeling capabilities. 
Compared to Hypersigma, a dedicated hyperspectral foundation model, our FFT SpectralGPT achieves better performance on Pavia, Houston, and Botswana datasets, with differences reaching +7.87\% OA for Houston. Conversely, Hypersigma outperforms FFT SpectralGPT on Indian Pines and WHU-Hi-LongKou, with OA differences of just +1.64\% and +1.9\% respectively. These results highlight the competitiveness of our method, as well as some other state-of-the-art models like MambaHSI, against a specialized hyperspectral foundation model. 

\subsubsection{Performance of PEFT Methods}

Starting with the linear probe (LP), which freezes the pre-trained weights and fine-tunes only the classifier, its limited performance underscores the necessity of unfreezing and fine-tuning either all or parts of the pre-trained weights. 
Bitfit, which fine-tunes only the biases of the targeted layers, achieves better performance than the linear probe. On Botswana, for example, Bitfit improves OA by +2\% over the linear probe. LoRA which adapts more weights than BitFit achieves better performance on most datasets, but BitFit slightly outperforms LoRA on Pavia and Houston, suggesting that LoRA rank optimization could improve results for these datasets. LoRA+, which applies differentiated learning rates to the LoRA adapter matrices, brings a small improvement (+0.5\% OA for example on Botswana) over LoRA, confirming the importance of asymmetric fine-tuning. KronA systematically outperforms LoRA-based methods, demonstrating for example +1.2\% OA gain over LoRA+ on Botswana. This is due to the fact that kronecker product structure increases expressiveness, bypassing LoRA’s low-rank constraint while keeping parameter efficiency. LoKr, which combines Kronecker decomposition and low-rank adaptation, shows better performance than LoRA+ on all datasets, but slightly underperforms KronA.  
Our proposed KronA+ method emerges as the best-performing PEFT approach, surpassing all others across datasets. It achieves an additional +1.5\% OA compared to KronA and comes remarkably close to FFT SpectralGPT, with a difference of just 0.3\% OA on Botswana. Moreover, KronA+ outperforms the SOTA backbones used for comparison in this study on Pavia and Houston datasets. These performances underscore the effectiveness of extending the differentiated learning rate strategy from LoRA+ to KronA.

\begin{table*}[ht]
\centering
\caption{Performance of the different classification methods on the Indian Pines dataset}
\label{table:IP}
\setlength{\tabcolsep}{3pt} 
\resizebox{\textwidth}{!}{%
\begin{tabular}{lccccccccccc}
    \toprule
    \textbf{Class} & \textbf{LP} & \textbf{BitFit} & \textbf{LoRA} & \textbf{LoRA+} & \textbf{LoKr} & \textbf{Krona} & \textbf{Krona+} & \textbf{FFT} & \textbf{Spectral-} & \textbf{MambaHSI} & \textbf{Hyper-} \\
    & & & & & &  & & \textbf{SpectralGPT}& \textbf{former} & & \textbf{SIGMA}\\
    \midrule
    1  & 52.17 & 51.73 & 67.43 & 64.41 & 69.64 & 60.35 & 68.31 & 79.96 & 74.13 & 92.49 & 79.59 \\
    2  & 0.00  & 50.89 & 71.25 & 64.11 & 66.61 & 72.14 & 80.71 & 77.86 & 88.01 & 95.41 & 62.71 \\
    3  & 4.38  & 98.54 & 93.43 & 97.81 & 99.27 & 97.81 & 96.35 & 97.81 & 88.59 & 100.00 & 75.68 \\
    4  & 0.00  & 26.67 & 80.58 & 69.28 & 88.41 & 68.41 & 84.64 & 86.96 & 95.97 & 96.20 & 89.86 \\
    5  & 61.41 & 92.97 & 90.10 & 90.82 & 91.39 & 93.11 & 87.52 & 95.98 & 84.65 & 99.43 & 95.55 \\
    6  & 78.00 & 90.75 & 94.25 & 93.25 & 98.75 & 98.00 & 98.25 & 99.50 & 94.76 & 99.77 & 99.46 \\
    7  & 24.09 & 69.83 & 77.37 & 83.21 & 77.98 & 82.36 & 83.45 & 80.66 & 77.34 & 90.74 & 90.79 \\
    8  & 7.50  & 76.93 & 73.69 & 76.31 & 79.33 & 81.19 & 79.15 & 78.57 & 80.85 & 96.20 & 82.74 \\
    9  & 0.21  & 62.19 & 75.83 & 79.34 & 71.28 & 79.34 & 78.93 & 81.40 & 75.89 & 90.60 & 87.94 \\
    10 & 90.74 & 100.0 & 100.0 & 100.0 & 100.0 & 100.0 & 100.0 & 100.0 & 99.38 & 100.00 & 100.00 \\
    11 & 92.03 & 97.51 & 94.85 & 95.68 & 93.19 & 96.35 & 93.11 & 97.59 & 93.65 & 98.79 & 98.14 \\
    12 & 0.00  & 86.96 & 88.82 & 83.23 & 91.30 & 90.06 & 93.17 & 96.27 & 84.85 & 99.70 & 100.00 \\
    13 & 100.0 & 93.33 & 97.78 & 93.33 & 95.56 & 100.0 & 97.78 & 100.0 & 100.00 & 100.00 & 100.00 \\
    14 & 0.00  & 74.36 & 94.87 & 89.74 & 82.05 & 87.18 & 89.74 & 84.62 & 74.36 & 97.44 & 100.00 \\
    15 & 0.00  & 100.0 & 90.91 & 100.0 & 90.91 & 100.0 & 100.0 & 100.0 & 100.00 & 100.00 & 100.00 \\
    16 & 0.00  & 100.0 & 100.0 & 100.0 & 100.0 & 100.0 & 100.0 & 100.0 & 100.00 & 100.00 & 100.00 \\
    \midrule
    \textbf{\#Params} & 0.012 & 0.031 & 0.16 & 0.16 & 0.051 & 0.049 & 0.049 & 85.266 & 0.356 & 0.425 & 182.632 \\
    \textbf{(x1e6)}   &       &       &       &       &       &       &       &       &   &   &   \\
    \midrule
    \textbf{OA (\%)}  & 35.96 & 73.67 & 79.83 & 79.94 & 81.61 & 81.62 & \textcolor{red}{\textbf{82.76}} & \textbf{85.62} & 83.79 & \textcolor{blue}{\textbf{95.77}} & 87.26 \\
\textbf{AA (\%)}  & 31.91 & 79.54 & 86.95 & 86.28 & 87.23 & 87.89 & \textcolor{red}{\textbf{89.44}} & \textbf{90.37} & 88.28 & \textcolor{blue}{\textbf{97.30}} & 91.40 \\
\textbf{Kappa (\%)} & 28.88 & 69.67 & 76.89 & 76.94 & 78.82 & 78.80 & \textcolor{red}{\textbf{80.19}} & \textbf{83.41} & 81.46 & \textcolor{blue}{\textbf{94.92}} & 84.91 \\
    \bottomrule
\end{tabular}%
}
\end{table*}

\begin{table*}[ht]
\centering
\caption{Performance of the different classification methods on Pavia dataset}
\setlength{\tabcolsep}{3pt} 
\resizebox{\textwidth}{!}{%
\begin{tabular}{lccccccccccc}
    \toprule
    \textbf{Class} & \textbf{LP} & \textbf{BitFit} & \textbf{LoRA} & \textbf{LoRA+} & \textbf{LoKr} & \textbf{Krona} & \textbf{Krona+} & \textbf{FFT} & \textbf{Spectral-} & \textbf{MambaHSI} & \textbf{Hyper-} \\
    & & & & & &  & & \textbf{SpectralGPT}& \textbf{former} & & \textbf{SIGMA}\\
\midrule
1  & 67.54 & 88.87 & 94.10 & 93.51 & 92.29 & 95.31 & 94.96 & 98.24 & 80.96 & 90.24 & 86.59 \\
2  & 60.53 & 90.25 & 93.62 & 94.82 & 90.19 & 94.13 & 95.88 & 94.28 & 97.52 & 93.84 & 94.93 \\
3  & 6.58  & 62.28 & 58.66 & 49.67 & 79.90 & 69.58 & 66.93 & 83.40 & 73.33 & 57.30 & 82.64 \\
4  & 90.44 & 94.35 & 93.19 & 91.21 & 93.97 & 93.40 & 92.48 & 97.11 & 97.66 & 90.93 & 58.40 \\
5  & 97.66 & 99.73 & 99.28 & 97.75 & 99.46 & 99.10 & 98.92 & 99.01 & 100.00 & 99.91 & 98.92 \\
6  & 41.40 & 89.06 & 61.88 & 75.22 & 87.14 & 81.58 & 78.52 & 88.89 & 65.31 & 83.03 & 66.01 \\
7  & 42.81 & 84.20 & 92.86 & 96.33 & 98.67 & 93.58 & 97.86 & 85.73 & 87.16 & 85.22 & 85.63 \\
8  & 92.93 & 94.77 & 90.70 & 94.92 & 93.70 & 96.85 & 96.82 & 91.62 & 95.10 & 98.48 & 88.03 \\
9  & 84.40 & 95.35 & 96.48 & 96.35 & 96.23 & 96.10 & 96.23 & 96.86 & 92.83 & 98.24 & 69.46 \\
\midrule
\textbf{\#Params} & 0.007 & 0.025 & 0.154 & 0.154 & 0.045 & 0.044 & 0.044 & 85.266 & 0.171 & 0.412 & 182.632 \\
\textbf{(x1e6)}   &       &       &       &       &       &       &       &       &   &   &   \\
\midrule
\textbf{OA (\%)}  & 63.22 & 89.60 & 88.27 & 90.18 & 90.92 & 91.68 & \textcolor{red}{\textbf{92.89}} & \textbf{93.74} & 89.66 & \textcolor{blue}{\textbf{90.61}} & 84.80 \\
\textbf{AA (\%)}  & 64.92 & 88.76 & 86.75 & 87.75 & 92.39 & 91.47 & \textcolor{red}{\textbf{90.13}} & \textbf{92.79} & 87.76 & \textcolor{blue}{\textbf{88.58}} & 81.18 \\
\textbf{Kappa (\%)} & 53.50 & 86.35 & 84.29 & 86.90 & 88.09 & 89.50 & \textcolor{red}{\textbf{90.53}} & \textbf{91.71} & 85.91 & \textcolor{blue}{\textbf{85.05}} & 80.27 \\
\bottomrule
\end{tabular}
}
\label{table:pavia}
\end{table*}

\begin{table*}[ht]
\centering
\caption{Performance of the different classification methods on Houston2013 dataset}
\setlength{\tabcolsep}{3pt} 
\resizebox{\textwidth}{!}{%
\begin{tabular}{lccccccccccc}
    \toprule
    \textbf{Class} & \textbf{LP} & \textbf{BitFit} & \textbf{LoRA} & \textbf{LoRA+} & \textbf{LoKr} & \textbf{Krona} & \textbf{Krona+} & \textbf{FFT} & \textbf{Spectral-} & \textbf{MambaHSI} & \textbf{Hyper-} \\
    & & & & & &  & & \textbf{SpectralGPT}& \textbf{former} & & \textbf{SIGMA}\\
    \midrule
    
1  & 82.05 & 82.05 & 82.15 & 80.63 & 78.44 & 81.86 & 81.58 & 81.77 & 82.24 & 82.34 & 82.25 \\
2  & 61.56 & 99.81 & 98.03 & 99.44 & 100.00 & 96.52 & 97.93 & 100.00 & 99.34 & 99.34 & 98.32 \\
3  & 98.42 & 99.80 & 98.81 & 98.61 & 97.23 & 99.41 & 99.41 & 99.41 & 91.68 & 72.48 & 100.00 \\
4  & 91.76 & 95.55 & 96.50 & 94.98 & 95.17 & 98.20 & 96.69 & 94.98 & 99.34 & 97.25 & 79.68 \\
5  & 96.78 & 98.77 & 99.91 & 98.86 & 97.16 & 98.67 & 99.72 & 99.81 & 100.00 & 100.00 & 93.83 \\
6  & 57.34 & 81.12 & 88.11 & 88.81 & 91.61 & 74.13 & 75.52 & 96.50 & 93.71 & 94.41 & 95.80 \\
7  & 60.60 & 97.36 & 92.55 & 97.46 & 91.89 & 91.05 & 91.52 & 93.40 & 93.75 & 75.37 & 80.04 \\
8  & 55.74 & 65.68 & 88.43 & 87.00 & 88.53 & 82.89 & 86.81 & 93.50 & 71.23 & 68.28 & 77.10 \\
9  & 51.85 & 80.63 & 85.38 & 85.09 & 73.31 & 83.76 & 78.44 & 84.33 & 78.28 & 91.88 & 84.34 \\
10 & 26.83 & 82.53 & 48.46 & 57.43 & 80.79 & 69.31 & 70.85 & 76.25 & 52.70 & 81.08 & 57.95 \\
11 & 64.10 & 85.81 & 93.14 & 87.05 & 88.48 & 82.48 & 88.38 & 82.00 & 86.53 & 87.10 & 73.18 \\
12 & 0.00  & 86.74 & 86.65 & 85.40 & 85.49 & 89.63 & 88.86 & 89.43 & 82.52 & 82.90 & 82.42 \\
13 & 62.11 & 87.72 & 91.23 & 89.47 & 86.67 & 89.12 & 92.28 & 89.12 & 75.79 & 82.11 & 77.01 \\
14 & 76.92 & 97.98 & 97.98 & 97.57 & 96.76 & 95.14 & 97.17 & 100.00 & 98.79 & 99.60 & 100.00 \\
15 & 81.82 & 100.00 & 99.58 & 99.58 & 100.00 & 98.31 & 99.58 & 100.00 & 94.50 & 98.52 & 94.50 \\
\midrule
\textbf{\#Params} & 0.012 & 0.03  & 0.159 & 0.159 & 0.05  & 0.048 & 0.048 & 85.265 & 0.236 & 0.418 & 182.632 \\
\textbf{(x1e6)}   &       &       &       &       &       &       &       &       &   &   &   \\
\midrule
\textbf{OA (\%)}  & 62.17 & 88.68 & 88.49 & 88.62 & 89.00 & 89.15 & \textcolor{red}{\textbf{89.59}} & \textbf{90.68} & 85.54 & \textcolor{blue}{86.70} & 82.81 \\
\textbf{AA (\%)}  & 64.53 & 89.44 & 89.79 & 89.83 & \textcolor{red}{\textbf{90.10}} & 89.78 & 89.38 & \textbf{92.03} & 86.69 & \textcolor{blue}{87.51} & 85.10 \\
\textbf{Kappa (\%)} & 59.20 & 87.71 & 87.50 & 87.64 & 88.07 & 88.22 & \textcolor{red}{\textbf{88.70}} & \textbf{89.89} & 84.29 & \textcolor{blue}{85.50} & 81.35 \\
\bottomrule
\end{tabular}
}
\label{table:houston}
\end{table*}

\begin{table*}[ht]
\centering
\caption{Performance of the different classification methods on Botswana dataset}
\setlength{\tabcolsep}{3pt} 
\resizebox{\textwidth}{!}{%
\begin{tabular}{lccccccccccc}
    \toprule
    \textbf{Class} & \textbf{LP} & \textbf{BitFit} & \textbf{LoRA} & \textbf{LoRA+} & \textbf{LoKr} & \textbf{Krona} & \textbf{Krona+} & \textbf{FFT} & \textbf{Spectral-} & \textbf{MambaHSI} & \textbf{Hyper-} \\
    & & & & & &  & & \textbf{SpectralGPT}& \textbf{former} & & \textbf{SIGMA}\\
    \midrule
1  & 100.00 & 99.58 & 98.33 & 98.75 & 99.58 & 98.33 & 99.58 & 99.58 & 100.00 & 100.00 & 95.83 \\
2  & 100.00 & 100.00 & 100.00 & 100.00 & 100.00 & 100.00 & 100.00 & 100.00 & 100.00 & 100.00 & 100.00 \\
3  & 58.37  & 98.19  & 100.00 & 100.00 & 100.00 & 99.55 & 100.00 & 100.00 & 99.55  & 100.00 & 99.55 \\
4  & 83.24  & 88.65  & 93.51  & 95.68  & 97.30  & 98.38 & 98.92  & 98.38  & 96.76  & 100.00 & 99.46 \\
5  & 11.30  & 89.54  & 96.23  & 94.56  & 99.16  & 97.07 & 96.23  & 99.16  & 81.59  & 82.01  & 98.33 \\
6  & 66.95  & 84.10  & 94.14  & 90.38  & 91.21  & 91.21 & 93.31  & 95.82  & 92.89  & 100.00 & 100.00 \\
7  & 83.41  & 100.00 & 97.38  & 99.56  & 93.01  & 98.25 & 98.25  & 100.00 & 99.56  & 100.00 & 99.56 \\
8  & 22.54  & 100.00 & 97.69  & 100.00 & 99.42  & 98.84 & 100.00 & 100.00 & 95.95  & 67.63  & 100.00 \\
9  & 22.18  & 93.31  & 95.42  & 98.59  & 97.18  & 98.94 & 98.59  & 98.59  & 96.48  & 97.54  & 98.94 \\
10 & 90.83  & 95.41  & 99.54  & 97.25  & 97.71  & 98.62 & 99.08  & 96.33  & 100.00 & 94.95  & 100.00 \\
11 & 91.27  & 98.18  & 95.64  & 96.36  & 98.91  & 98.18 & 98.91  & 100.00 & 98.91  & 97.45  & 98.10 \\
12 & 100.00 & 99.33  & 100.00 & 100.00 & 100.00 & 100.00 & 100.00 & 99.33  & 100.00 & 98.68  & 95.24 \\
13 & 88.66  & 100.00 & 99.58  & 99.58  & 100.00 & 100.00 & 100.00 & 100.00 & 100.00 & 99.58  & 95.88 \\
14 & 81.54  & 100.00 & 100.00 & 100.00 & 100.00 & 100.00 & 100.00 & 100.00 & 100.00 & 100.00 & 100.00 \\
\midrule
\textbf{\#Params} & 0.011  & 0.029  & 0.158  & 0.158  & 0.049  & 0.048  & 0.048  & 85.264 & 0.237  & 0.418  & 182.632  \\
\textbf{(x1e6)}   &        &        &        &        &        &        &        &       &    &    &    \\
\midrule
\textbf{OA (\%)}  & 68.51  & 95.61  & 97.31  & 97.56  & 97.81  & 98.12 & \textcolor{red}{98.55}  & \textbf{98.97} & 96.85  & 95.51  & \textcolor{blue}{98.65} \\
\textbf{AA (\%)}  & 71.45  & 96.16  & 97.68  & 97.91  & 98.11  & 98.39 & \textcolor{red}{98.78}  & \textbf{99.09} & 97.26  & 95.56  & \textcolor{blue}{98.63} \\
\textbf{Kappa (\%)} & 66.01  & 95.24  & 97.08  & 97.35  & 97.62  & 97.96 & \textcolor{red}{98.43}  & \textbf{98.89} & 96.58  & 94.51  & \textcolor{blue}{98.53} \\
\bottomrule
\end{tabular}
}
\label{table:botswana}
\end{table*}

\begin{table*}[ht]
\centering
\caption{Performance of the different classification methods on Whu-HI Longkou dataset}
\setlength{\tabcolsep}{3pt} 
\resizebox{\textwidth}{!}{%
\begin{tabular}{lccccccccccc}
    \toprule
    \textbf{Class} & \textbf{LP} & \textbf{BitFit} & \textbf{LoRA} & \textbf{LoRA+} & \textbf{LoKr} & \textbf{Krona} & \textbf{Krona+} & \textbf{FFT} & \textbf{Spectral-} & \textbf{MambaHSI} & \textbf{Hyper-} \\
    & & & & & &  & & \textbf{SpectralGPT}& \textbf{former} & & \textbf{SIGMA}\\
\midrule
1  & 0.00  & 90.41  & 93.36  & 90.10  & 92.40  & 93.93  & 90.45  & 98.03  & 99.70  & 99.77  & 99.73  \\
2  & 0.00  & 97.44  & 99.34  & 97.63  & 97.91  & 97.28  & 98.62  & 98.68  & 98.98  & 100.00 & 99.66  \\
3  & 0.00  & 87.24  & 94.57  & 97.99  & 94.09  & 96.04  & 95.75  & 99.29  & 97.64  & 99.90  & 100.00 \\
4  & 82.64 & 83.98  & 92.63  & 91.27  & 94.69  & 91.62  & 94.82  & 95.13  & 99.21  & 99.43  & 98.96  \\
5  & 11.50 & 85.48  & 95.74  & 97.30  & 90.02  & 93.47  & 91.59  & 96.31  & 89.43  & 99.78  & 99.89  \\
6  & 0.02  & 93.83  & 94.93  & 97.50  & 97.45  & 97.41  & 97.01  & 97.58  & 99.64  & 99.85  & 99.79  \\
7  & 41.06 & 93.29  & 96.85  & 96.88  & 95.04  & 99.10  & 98.30  & 99.23  & 99.96  & 99.75  & 99.64  \\
8  & 30.32 & 88.69  & 93.85  & 79.64  & 95.20  & 93.44  & 96.63  & 97.57  & 97.19  & 98.47  & 98.98  \\
9  & 41.20 & 90.88  & 92.88  & 95.15  & 94.62  & 97.78  & 95.90  & 98.08  & 98.96  & 99.24  & 99.82  \\
\midrule
\textbf{\#Params} & 0.007 & 0.025 & 0.154 & 0.154 & 0.045 & 0.044 & 0.044 & 85.26 & 0.558 & 0.433 & 182.632 \\
\textbf{(x1e6)}   &       &       &       &       &       &       &       &       &   &   &   \\
\midrule
\textbf{OA (\%)}  & 42.16 & 89.80 & 94.77 & 93.58 & 94.67 & 95.52 & \textcolor{red}{\textbf{95.71}} & \textbf{97.55} & 99.32 & \textcolor{blue}{99.62} & 99.45 \\
\textbf{AA (\%)}  & 22.97 & 90.14 & 94.90 & 93.72 & 94.60 & \textcolor{red}{\textbf{95.56}} & 95.45 & \textbf{97.77} & 97.86 & 99.58 & \textcolor{blue}{99.61} \\
\textbf{Kappa (\%)} & 21.41 & 86.62 & 93.05 & 91.46 & 92.93 & 94.02 & \textcolor{red}{\textbf{94.27}} & \textbf{96.73} & 99.09 & \textcolor{blue}{99.42} & 99.26 \\
\bottomrule
\end{tabular}
}
\label{table:longkou}
\end{table*}

\newpage

\subsection{Visual Analysis}

\begin{figure*}[ht]
    \centering
    \includegraphics[scale=0.1]{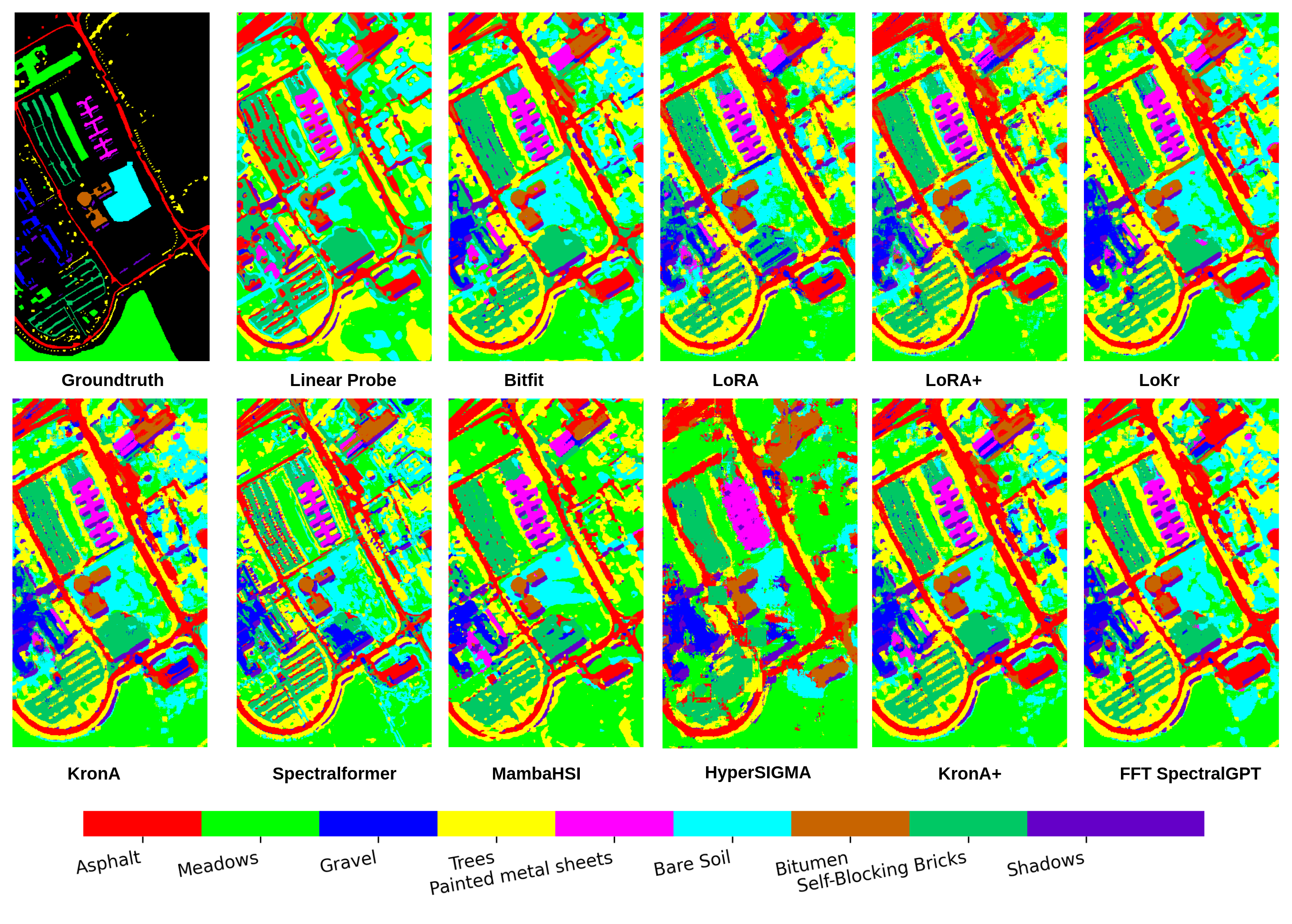}
    \caption{
        Predicted classification maps from various methods on the Pavia dataset.
    }
    \label{fig:pavia_maps}
\end{figure*}

\begin{figure*}[h!]
    \centering
    \includegraphics[scale=0.08]{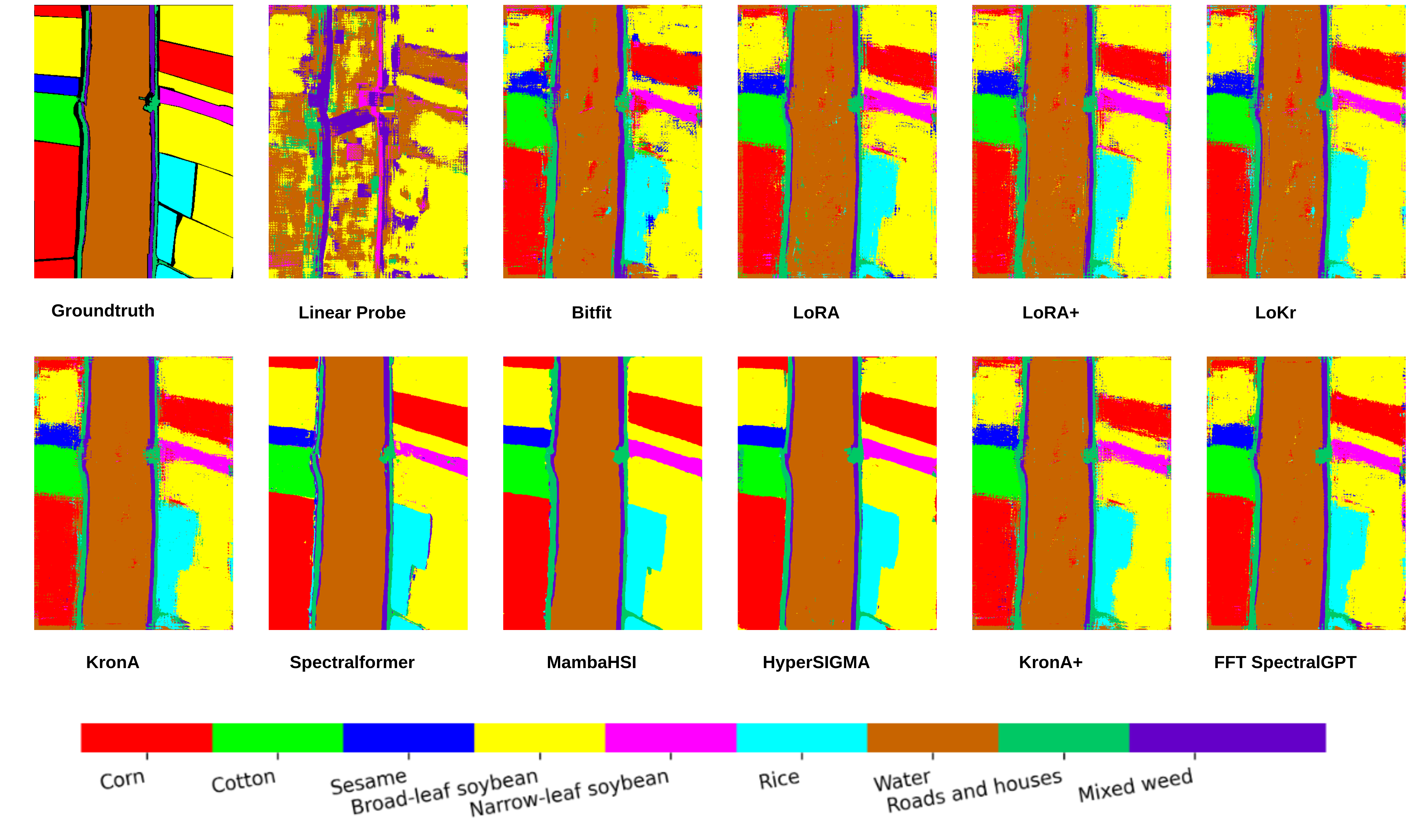}
    \caption{
        Predicted classification maps from various methods on the WHU-Hi LongKou dataset. 
    }
    \label{fig:longkou_maps}
\end{figure*}

The figures \ref{fig:pavia_maps} and \ref{fig:longkou_maps} provide a visual comparison of the classification maps generated by the different methods for Pavia and Whu-hi-Longkou datasets. As expected, the visual quality of these maps is correlated with the quantitative performance of the methods. Unsurprisingly, the linear probe produces the least accurate maps, with certain classes missing entirely, particularly on the Longkou dataset. On Pavia, our FFT SpectralGPT and the KronA+ method deliver visually comparable results for each dataset, with FFT SpectralGPT exhibiting slightly less noise than KronA+. Both methods achieve superior boundary delineation between classes compared to other PEFT approaches. On Longkou, where FFT SpectralGPT and KronA+ are not the top-performing methods, MambaHSI produces the most visually accurate maps with reduced noise and better class distinction, showing its robustness on this dataset.


\subsection{Number of Parameters Analysis}

The number of trainable parameters varies significantly between fine-tuning approaches, affecting performance. Linear Probe (LP) has the smallest parameter count and also the lowest performance, highlighting the limitations of freezing the backbone. A step further, BitFit doubles the parameters compared to LP, leading to a performance improvement, although it remains inferior to structured PEFT methods. LoRA and LoRA+ introduce approximately 0.150M parameters each, significantly more than BitFit, leading to notable performance gains. In contrast, KronA employs Kronecker factorization to reduce parameters ($\sim 0.05M$) while outperforming LoRA, confirming that the Kronecker decomposition improves expressiveness without increasing model size. Meanwhile, LoKr has a parameter count similar to KronA, achieves better performance than LoRA, but remains slightly behind KronA. The best-performing PEFT method, KronA+, maintains the same lightweight parameter footprint as KronA while outperforming all other PEFT methods due to the introduction of differentiated learning rates. Moving to full fine-tuning, FFT SpectralGPT requires $\sim 85.266 M$ parameters, leading to the highest OA performance metrics, confirming that fully adapting the model provides an advantage over PEFT. Among SOTA models, Spectralformer and MambaHSI require more parameters than PEFT methods,  but remain significantly smaller than FFT SpectralGPT. In contrast, Hypersigma is by far the largest model with $\sim 183 M$ parameters, yet it does not systematically outperform FFT SpectralGPT, reinforcing that larger models do not necessarily yield better performance. Additionally, FFT SpectralGPT and all PEFT methods are trained in just 50 epochs, whereas SOTA models require at least 100 epochs, and particularly Hypersigma needs 200 epochs, yet FFT SpectralGPT remains highly competitive, demonstrating the effectiveness of our approach.

\subsection{Adapters Storage}
In this study, in addition to the number of trainable parameters, we monitor the Adapters Storage (AS). In fact, in full fine-tuning, all model parameters must be updated and stored, resulting in significant storage requirements. In contrast, PEFT methods add only adaptation-specific parameters (e.g., low-rank matrices in LoRA or bias terms in BitFit) while keeping the pre-trained model parameters frozen. This metric (AS) highlights the storage efficiency achieved by focusing solely on storing these lightweight adapters, which can be fused with the original pre-trained model during inference.

\begin{figure}[ht]
    \centering
    \includegraphics[width=0.8\textwidth]{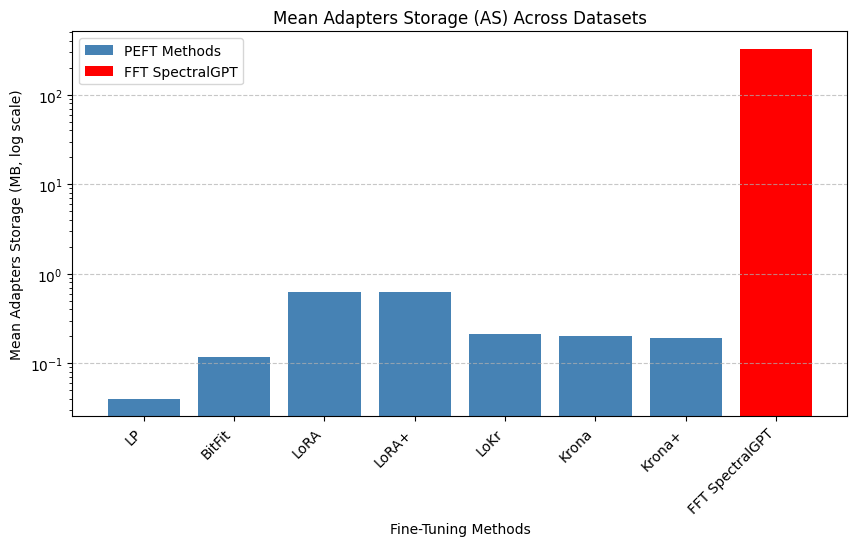} 
    \caption{ Mean Adapters Storage (AS) across fine-tuning methods. The y-axis is displayed in logarithmic scale to better highlight differences among methods.}
    \label{fig:mean_storage}
\end{figure}

To provide a clear comparison of the storage requirements across fine-tuning methods, we report the mean Adapters Storage (AS) across datasets in Table \ref{table:mean_storage} and visualize it in Figure \ref{fig:mean_storage}. The logarithmic scale in the figure highlights the differences among methods more effectively. As expected, LP and BitFit have the smallest storage footprints, while LoRA-based and Kronecker-based methods require slightly more storage. LoRA and LoRA+ have similar storage consumption ($\sim0.62$ MB), whereas LoKr, KronA, and KronA+ maintain lower memory costs ($\sim0.20$ MB). Our KronA+, the best PEFT method in terms of performance metrics, emerges here also among the most storage-efficient alternatives. In contrast, FFT SpectralGPT requires 326 MB, significantly larger than any PEFT method, reinforcing the advantage of parameter-efficient approaches in reducing storage overhead.

\begin{table*}[ht]
\centering
\caption{Mean Adapters Storage (AS) in MB across fine-tuning methods.}
\resizebox{\textwidth}{!}{%
\begin{tabular}{lcccccccc}
\toprule
\textbf{Method} & \textbf{LP} & \textbf{BitFit} & \textbf{LoRA} & \textbf{LoRA+} & \textbf{LoKr} & \textbf{Krona} & \textbf{Krona+} & \textbf{FFT SpectralGPT} \\
\midrule
Mean AS (MB) & 0.04 & 0.12 & 0.62 & 0.62 & 0.22 & 0.20 & 0.20 & 326 \\
\bottomrule
\end{tabular}
}
\label{table:mean_storage}
\end{table*}

\subsection{The Cost-Performance Trade-off in Hyperspectral Foundation Models}
A key observation from our experiments is that no single model consistently dominates across all datasets, highlighting the complexity and variability inherent to hyperspectral images. While certain performance patterns emerge, they do not hold universally, as demonstrated by WHU-Hi-LongKou, where all SOTA backbones outperform FFT SpectralGPT by $\sim2\%$ OA, despite our model surpassing them on nearly all other datasets. 

A particularly striking result is the performance of Hypersigma, a domain-specific hyperspectral foundation model. Despite its large-scale pretraining and fine-tuning over 200 epochs—four times more than FFT SpectralGPT (50 epochs)—its performance remains comparable to, or sometimes even lower than, that of well-optimized transformer or state-space models trained on significantly fewer epochs. This aligns with the findings in \textit{PANGAEA: A Global and Inclusive Benchmark for Geospatial Foundation Models}  \citep{pangea}, which question the assumption that a domain-specific foundation model necessarily guarantees superior performance over smaller, well-fine-tuned architectures. Similarly, \textit{Foundation Models - A Panacea for Artificial Intelligence in Pathology?}  \citep{pathology_fm} highlights that large-scale pathology foundation models, despite their extensive pretraining, do not consistently outperform task-specific models and often come with significantly higher computational costs. These examples reinforce that across multiple domains—including remote sensing and pathology—foundation models, despite their large number of parameters, do not always guarantee superior performance, highlighting the importance of evaluating cost-performance trade-offs in specialized applications.


\subsection{Ablation Study}

In this section, we perform a series of experiments to analyze the effect of various training strategies in both full fine-tuning and PEFT  methods using the Botswana and Pavia datasets. The impact of each experiment is quantitatively assessed using Overall Accuracy (OA) and the Kappa coefficient.

\subsubsection{Effect of Token Size}

In the Vision Transformer architecture, the input is divided into non-overlapping tokens with a specified patch size. We explore the impact of varying the token patch size to understand its influence on model performance.

\begin{table}[h]
\centering
\caption{Effect of token patch size on OA and Kappa for the Pavia and Botswana datasets.}
\begin{tabular*}{\textwidth}{@{\extracolsep\fill}lcccc}
\toprule%
& \multicolumn{2}{@{}c@{}}{Pavia Dataset} & \multicolumn{2}{@{}c@{}}{Botswana Dataset} \\\cmidrule{2-3}\cmidrule{4-5}%
Token Size & 8 x 8 & 16 x 16 & 8 x 8  &  16 x 16 \\
\midrule
OA (\%) & 93.21  & 91.67 & 98.90 & 98.30 \\
Kappa (\%)  & 90.97 & 88.93   & 98.81 & 98.16 \\
\botrule
\end{tabular*}
\label{tab:token_patch_size_effect}
\end{table}

As shown in Table~\ref{tab:token_patch_size_effect}, we observe that a token size of 8×8 yields better performance compared to 16×16. The OA  and Kappa values are higher for the smaller token size, highlighting its advantage in capturing the spatial-spectral representation of the inputs. These results are consistent with the findings reported in the official SpectralGPT paper  \citep{spectralgpt}.

\subsubsection{Effect of Embedding Dimension}

The Vision Transformer offers different configurations based on embedding dimensions: vit\_base (768), vit\_large(1024), and vit\_huge(1280), each with a progressively larger number of parameters. We evaluate the impact of varying the embedding dimension on model performance.

\begin{table}[h]
\centering
\caption{Effect of embedding dimension on OA and Kappa for the Pavia and Botswana datasets.}
\begin{tabular*}{\textwidth}{@{\extracolsep\fill}lcccccc}
\toprule%
& \multicolumn{3}{@{}c@{}}{Pavia Dataset} & \multicolumn{3}{@{}c@{}}{Botswana Dataset} \\\cmidrule{2-4}\cmidrule{5-7}%
ViT Size & Base & Large & Huge  &  Base & Large & Huge \\
\midrule
Params & 86M & 307M & 632M  &   86M & 307M & 632M \\
OA (\%) & 93.21  & 91.37 & 89.78 & 98.90 & 98.27& 98.55\\
Kappa (\%)  & 90.97       & 88.57       & 86.54  & 98.81 & 98.12        & 98.43 \\
\botrule
\end{tabular*}
\label{tab:embedding_dimension_effect}
\end{table}

The table \ref{tab:embedding_dimension_effect} shows that the \texttt{vit\_base} configuration, with the smallest number of parameters (86M), outperforms the larger configurations in terms of OA and Kappa, achieving the best results. This observation underscores that a smaller, more efficient model can be optimal for hyperspectral image classification tasks, where excessive model complexity does not necessarily translate to better performance. This finding further highlights the pertinence of employing parameter-efficient fine-tuning methods, which leverage lightweight configurations to achieve competitive performance while reducing computational and storage overheads.

\subsubsection{Study of lambda parameter in LoRA+ and KronA+}

Here, the goal is to study the optimal \(\lambda\) values where LoRA+ outperforms simple LoRA and KronA+ outperforms simple KronA.

\begin{figure}[ht]
    \centering
        \includegraphics[width=0.9\textwidth, height=7cm]
        {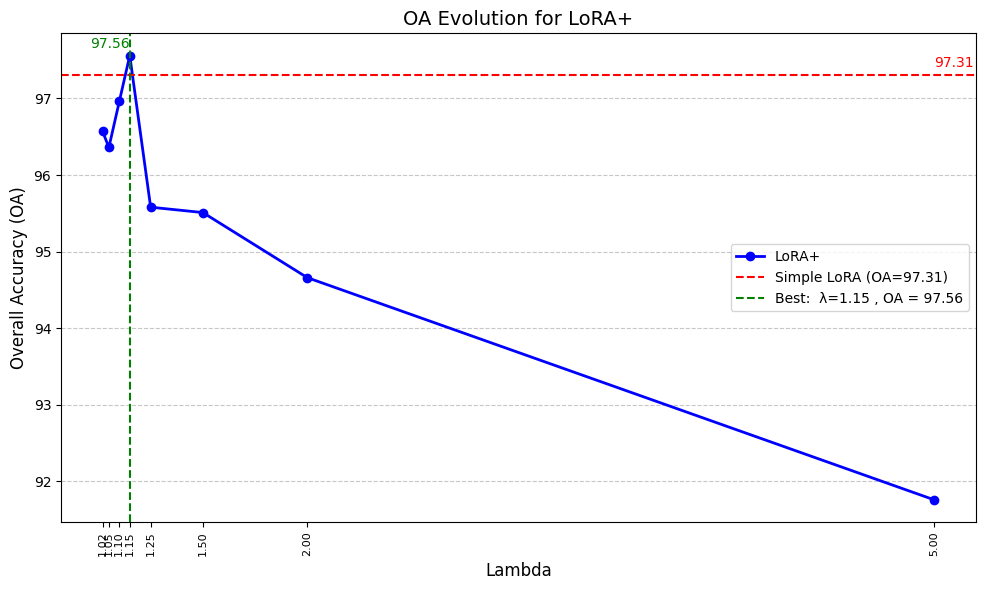}
        \caption{OA of LoRA+ with varying $\lambda$ for Botswana Dataset}
        \label{fig:lora_trend_botswana}
\end{figure}

\begin{figure}[ht]
    \centering
        \includegraphics[width=0.9\textwidth, height=7cm]{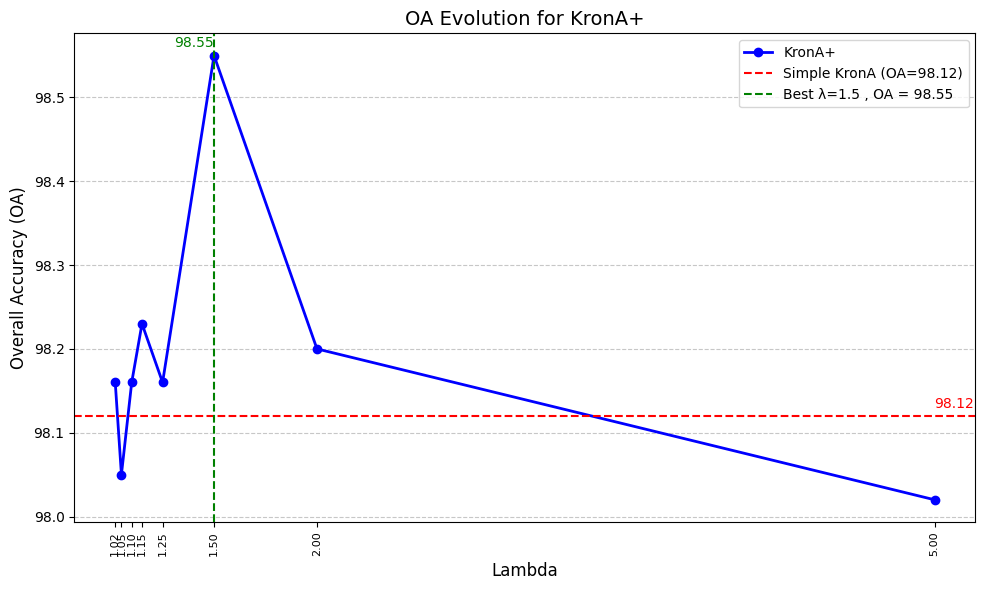}
        \caption{OA of KronA+ with varying $\lambda$ for Botswana Dataset}
        \label{fig:krona_trend_botswana}
    \label{fig:lora_krona_lambda_botswana}
\end{figure}

\begin{figure}[h]
    \centering
      \includegraphics[width=0.9\textwidth, height=7cm]{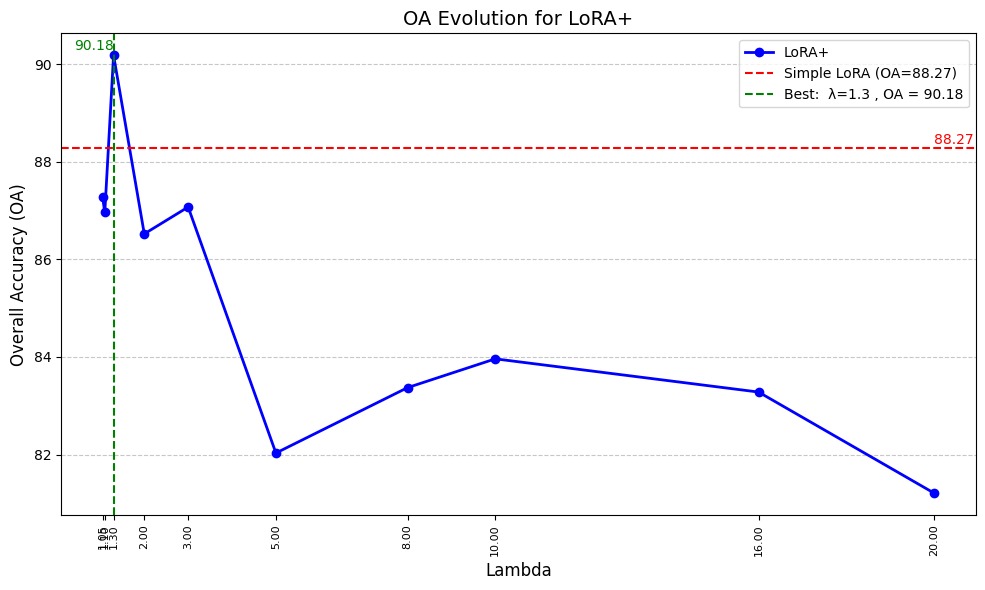}
        \caption{OA of LoRA+ with varying $\lambda$ for Pavia Dataset}
        \label{fig:lora_trend_pavia}
\end{figure}

\begin{figure}[h]
        \centering
        \includegraphics[width=0.9\textwidth, height=7cm]{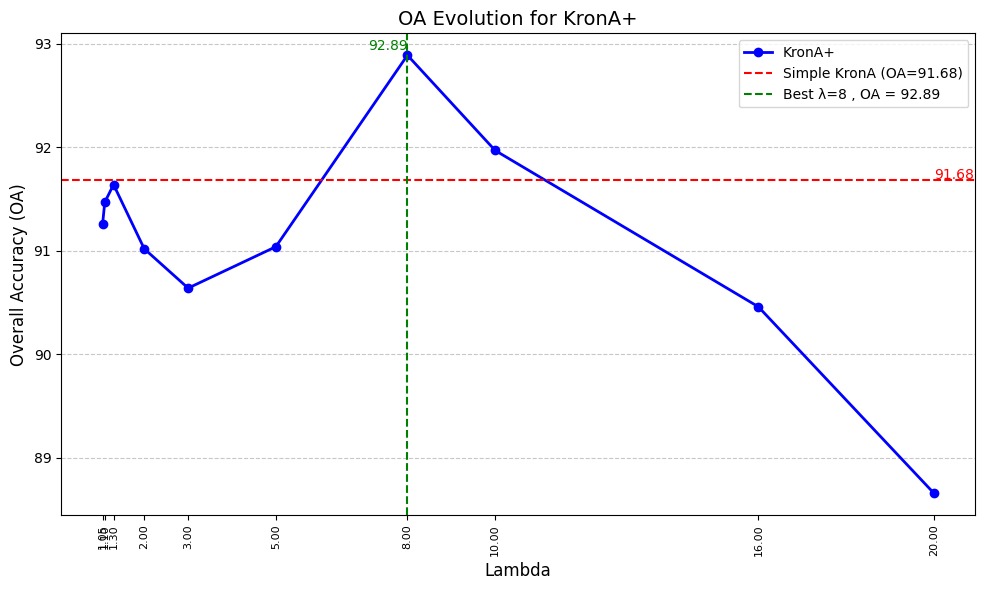}
        \caption{OA of KronA+ with varying $\lambda$ for Pavia Dataset}
        \label{fig:krona_trend_pavia}
    \label{fig:lora_krona_lambda_pavia}
\end{figure}

\begin{table}[h]
\centering
\small
\caption{\(\lambda\) values used for LoRA+ and KronA+ across different datasets.}
\begin{tabular}{lcccccc}
\hline
 & \textbf{Indian Pines} & \textbf{Pavia} & \textbf{Houston} & \textbf{Botswana} & \textbf{Longkou} \\ \hline
\textbf{LoRA+}          & 1.025                 & 1.3            & 1.15             & 1.15              & 1.08             \\ \hline
\textbf{KronA+}         & 1.02                  & 8           & 1.05             & 1.5               & 1.02             \\ \hline
\end{tabular}
\label{tab:lambda_values}
\end{table}

Figures \ref{fig:lora_krona_lambda_botswana} and \ref{fig:lora_krona_lambda_pavia} show OA values with varying \(\lambda\). For Botswana, LoRA+ achieves an OA of 97.56\% at \(\lambda = 1.15\), outperforming simple LoRA (OA = 97.31\%), while KronA+ peaks at \(\lambda = 1.5\), surpassing simple KronA (OA = 98.12\%). These results underscore the advantage of introducing \(\lambda\), which scales the learning rates of the low-rank matrices \(A\) and \(B\), enabling better adaptation.

Analyzing the values of \(\lambda\) in the HSI datasets (Table \ref{tab:lambda_values}), we observe that all optimal values for LoRA+ remain close to 1, showing that a small increase of \(\eta_B\) relative to \(\eta_A\) is sufficient to improve performance over the standard LoRA method. This behavior contrasts with text-based tasks, where a much larger scaling factor (\(\lambda \sim 2^4\) for Roberta-base finetuned on the MNLI
task) is often required to achieve optimal performance, as reported in the official LoRA+ paper \citep{lora+}. However, for KronA+, the Pavia dataset exhibits a significantly larger \(\lambda\) (8), suggesting that the optimal scaling factor variation depends on the dataset's characteristics. 
These findings align with a key limitation discussed in the LoRA+ paper: \textit{the optimal ratio \(\eta_B / \eta_A\) is both task- and model-dependent}. While our results suggest that the optimal value of $\lambda$ is generally close to 1 for HSI classification tasks, the case of KronA+ on the Pavia dataset highlights the need for a deeper exploration of how dataset-specific intrinsic characteristics influence the choice of $\lambda$, as similarly suggested in  \citep{lora+}.

\section{Conclusion}

In this paper, we present an effective framework to fine-tune, a multispectral foundation model (here SpectralGPT), for hyperspectral image classification tasks. Our experiments across multiple datasets demonstrate the ability of our framework to compete with state-of-the-art models in the hyperspectral imaging domain. By leveraging Parameter-Efficient Fine-Tuning (PEFT) methods, such as LoRA, LoRA+, LoKr, KronA, and the newly introduced KronA+, we successfully made the model lightweight and memory-efficient, enabling effective fine-tuning even in resource-constrained environments. Among the tested PEFT methods, KronA+ proved to be the most effective, benefiting from the distinct learning rate strategy of LoRA+ and outperforming other approaches, further validating its potential in adapting large-scale models to hyperspectral tasks. Additionally, our findings highlight that hyperspectral-specific foundation models like Hypersigma, despite extensive pretraining, do not necessarily outperform well-optimized transformer and state-space models, echoing similar discussions in other domains regarding the cost-performance trade-off of foundation models. 
Despite its contributions, this study has several limitations. The PEFT methods compared in this study are limited to partial-based tuning \citep{peft_survey} approaches. A broader perspective could be gained by exploring PEFT methods from other categories, such as adapter tuning \citep{adapter}, which could provide additional opportunities for improvement. Additionally, while KronA+ has demonstrated strong performance in hyperspectral image classification, the choice of the optimal lambda ratio in the hyperspectral domain remains an open question. A deeper theoretical study on how to best determine this scaling factor could further refine the method and improve adaptation strategies. Furthermore, in an era where multimodal foundation models \citep{oneforall,skysense} capable of handling RGB, multispectral, and hyperspectral data are emerging, future comparisons could include these models and our FFT SpectralGPT to offer a more comprehensive evaluation and contribute to a broader understanding of the field.

\section*{Data and Code Availability}
The data and code of this work will be available at\\
\url{https://github.com/LiganiumInc/PEFT\_HSIC} for the sake of reproducibility.

\bibliography{main}

\end{document}